\documentclass[pdftex,11pt,openright,headsepline]{book}
\usepackage{float}
\usepackage{pifont}
\usepackage{paralist}		
\usepackage{color}		
\usepackage{times}
\usepackage{amsfonts}		
\usepackage{amsmath}		
\usepackage{latexsym}
\usepackage{graphicx}		
\usepackage{mydefs}		
\usepackage{fancyhdr}		
\usepackage{fullpage} 
\usepackage{hyperref}

\usepackage[headsep=1cm]{geometry}
\usepackage{indentfirst}
\usepackage{comment}
\newcommand{\R}{\mathbb{R}}
\newcommand{\E}{\mathbb{E}}
\newcommand{\D}{\mathcal{D}}
\newcommand{\G}{\mathcal{G}}
\newcommand{\x}{\mathbf{x}}
\newcommand{\z}{\mathbf{z}}

\newcommand{\f}{\mathbf{f}}
\newcommand{\bm}[1]{\mathbf{#1}}
\usepackage{setspace}
\usepackage{caption}
\begin{document}

\fancypagestyle{plain}{
  \renewcommand{\headrulewidth}{0.0pt}
  \fancyfoot{}
  \fancyhead{}
}

%

\begin{titlepage}

\thispagestyle{empty}

\fancypagestyle{empty}{
\lhead{\includegraphics[height=1.5cm]{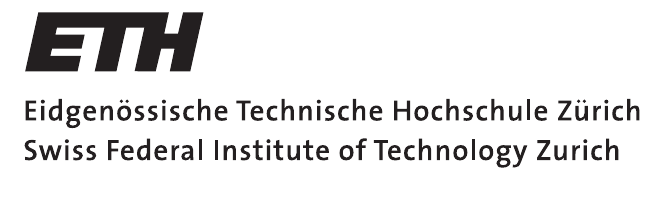}}
\renewcommand{\headrulewidth}{0.0pt}
\rhead{\vspace*{-0.2cm}\includegraphics[height=1.4cm]{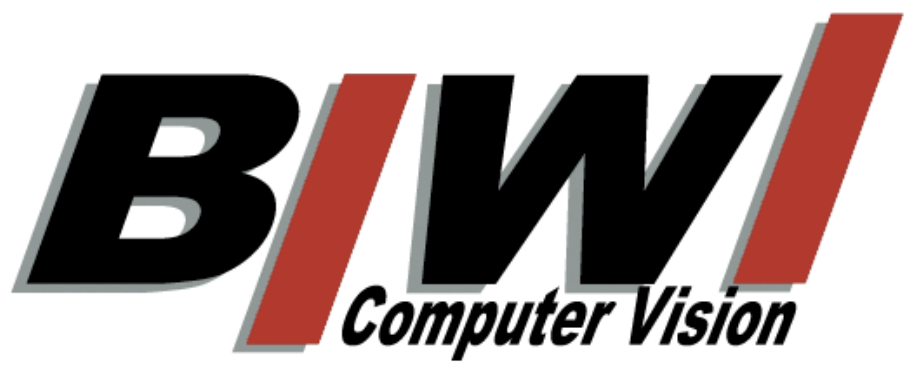}}
\fancyfoot{}
}

\vspace*{2cm}
\begin{center}
\Huge{\textbf{Towards High Resolution Video Generation with Progressive Growing of Sliced Wasserstein GANs
}\\}
\LARGE{\textbf{}\\[1cm]}

\LARGE{Dinesh Acharya\\}
\normalsize{Department of Mathematics}
\end{center}

\begin{center}


\end{center}

\vfill
\begin{center}
\begin{tabular}{ll}
\Large{\textbf Advisor:} & \Large{Dr. Zhiwu Huang, Dr. Danda Paudel}\\
\Large{\textbf Supervisor:} & \Large{Prof.~Dr.~Luc van Gool}\\
 			    & \small{Computer Vision Laboratory}\\
 			    & \small{Department of Information Technology and Electrical Engineering}\\
\end{tabular}
\end{center}

\begin{center}
May 22, 2018
\end{center}

\end{titlepage}

\cleardoublepage

%

\newpage
\vspace{3cm}

\chapter*{Abstract}
The extension of image generation to video generation turns out to be a very difficult task, since the temporal dimension of videos introduces an extra challenge during the generation process. Besides, due to the limitation of memory and training stability, the generation becomes increasingly challenging with the increase of the resolution/duration of videos. In this work, we exploit the idea of progressive growing of Generative Adversarial Networks (GANs) for higher resolution video generation. In particular, we begin to produce video samples of low-resolution and short-duration, and then progressively increase both resolution and duration alone (or jointly) by adding new spatiotemporal convolutional layers to the current networks. Starting from the learning on a very raw-level spatial appearance and temporal movement of the video distribution, the proposed progressive method learns spatiotemporal information incrementally to generate higher resolution videos. Furthermore, we introduce a sliced version of Wasserstein GAN (SWGAN) loss to improve the distribution learning on the video data of high-dimension and mixed-spatiotemporal distribution. SWGAN loss replaces the distance between joint distributions by that of one-dimensional marginal distributions, making the loss easier to compute. 

As suitable larger resolution video datasets are scarce, we collected 10,900 videos capturing face dynamics from Hollywood movie trailers. We used our new dataset to generate face videos of 256x256x32 resolution, while benefiting from sliced Wasserstein GAN loss within the progressive framework. Up to our knowledge, this is the first work that generates videos larger than 64x64x32 resolution. In addition to the gain on resolution, our model performs better than the existing methods in terms of both appearance and dynamics. The proposed model reaches a record inception score of 14.57 in unsupervised action recognition dataset UCF-101. Additionally, our method obtains a better FID score than the state-of-the-art methods, on two challenging datasets captured in the wild.

%

\newpage

\chapter*{Acknowledgements}
\noindent I am highly indebted to Dr. Zhiwu Huang and Dr. Danda Paudel for their supervision and support during the thesis. Content related to Sliced Wasserstein Gan in section 5 was taken from manuscript prepared by Dr. Zhiwu Huang. I would also like to thank Jiqing Wu for his code related to Sliced Wasserstein GAN. Prof. Dr. Luc Van Gool and his Computer Vision Lab also deserve special acknowledgement for the resources. I would also like to thank Bernhard Kratzwald whose models for TGAN and VideoGAN I used for some experiments.\\

\noindent I would also like to thank NVIDIA team for releasing their work on Progressive Growing of GANs and donating GPUs to the lab.\\

\noindent My parents and siblings deserve special acknowledgement for their help and support throughout all my endeavors.
\cleardoublepage
\newpage

\fancypagestyle{plain}{
  \renewcommand{\headrulewidth}{0.0pt}
  \fancyfoot{}
  \fancyfoot[RO, LE]{\thepage}
  \fancyhead{}
}

\pagestyle{fancy}
\pagenumbering{Roman}
\tableofcontents

\listoffigures
\cleardoublepage

\listoftables
\cleardoublepage

\newpage

\pagenumbering{arabic}

%

\chapter{Introduction}
Compared to images, videos contain additional temporal dynamics. Hence, richer information about the scene can be extracted. Furthermore, images are also restricted to a single perspective and are prone to ambiguity. Despite this, most of the focus of computer vision community has been on static image based techniques. This can be attributed to computational and storage overhead of video based algorithms as well as complexity of modeling videos. As seen from action recognition problems, static images are not sufficient to correctly predict the event. This holds true even when best possible frame is chosen~\cite{huangmakes}. This motivates the work on video based computer vision algorithms.

The main motivation behind generative models for both images and videos lies in the notion that how well we understand a given object is directly related to how well we can generate that object. As such, neural network based unsupervised methods such as Autoencoders~\cite{draw}\cite{vae1}, Autoregressive models~\cite{pixelrnn}\cite{pixelcnn}\cite{vpn} and Generative Adversarial Networks (GAN)~\cite{ganivan}\cite{iwgan}\cite{tgan} have been widely used to design generative models. Several of such techniques have also been applied to video generative models. These generative models provide mechanism for unsupervised representation learning and have been widely employed for semi-supervised methods. Semi-supervised and unsupervised techniques are particularly useful when collecting data is relatively easy but labelling is very expensive. Such scenarios arise in several problems such as semantic segmentation, medical image analysis and facial action unit encoding.

GANs are one of most widely used generative models. In last couple of years, research in GAN has made significant progress in terms of stability of training and quantitative evaluation frameworks~\cite{itgan}\cite{ttur}. These improvements can primarily be attributed to improved loss function~\cite{lsgan}\cite{wgan}\cite{iwgan}\cite{swgan}\cite{gmswd}, robust network architectures~\cite{dcgan}\cite{pggan} and robust training schemes~\cite{pggan} that guarantee convergence. However as in the case of other computer vision works, most of the GANs have focused on image based problems. The research into GANs that focus on video generation is relatively scarce. Due to computational limitation and network instability, all existing works generate $64\times 64$ tiny clips~\cite{mocogan}\cite{tgan}\cite{vgan}\cite{ftgan}.

While designing models for video generation, it is natural to assume that temporal variations in videos behave differently than spatial variations. Spatial variations can be used to infer various objects present in the scene, where as the dynamics of these objects can be inferred from temporal variations. The relationship between and interaction of these objects may depend on both temporal and spatial variations. So, while modeling videos, it is natural to model temporal and spatial domains separately. This has been explored by using 1-D convolutions for temporal generator \cite{tgan} or Recurrent Neural Networks (RNN) to generate latent code for image based generators \cite{mocogan}. Using 1-D convolutions also reduces the model size\cite{1d1}\cite{1d2}. However, 3-D convolutions have survived the test of time and are widely used for problems ranging from object recognition, shot detection to video stabilization.

In this work, we explore whether the robust measures introduced for training GANs on image based problems \cite{pggan}\cite{swgan} generalize to video based frameworks. In particular, we generalize the scheme of Progressive Growing of GANs\cite{pggan} to video based problems. As discussed earlier, though 1-D convolutions \cite{tgan} or Recurrent Neural Networks \cite{mocogan} have been experimented in the context of generative models to model the temporal domain, we use simple 3-D convolutions to keep the model simple. This is particularly important in light of the complexity of the model introduced by progressive growing approach.

Realistic video generation has been accomplished in works such as~\cite{deepvideoportrait}\cite{lipsinc}\cite{timelapse}\cite{videoprediction}\cite{lecun}. However these works are restricted to very specific problems and domains. As such, they are not useful for general unsupervised representation learning or for use in semi-supervised techniques. Furthermore, the network architectures in such scenarios are highly specialized. In this work, we explore the more general problem of unsupervised video generation using Generative Adversarial Networks. In particular, we apply the idea of Progressively Growing GANs for video generation.

\section{Focus of this Work}
The main focus of the work is on unsupervised generation of higher resolution videos. Such an endeavour has several challenges. First, there is lack of sufficient large resolution video datasets. Next, generating larger videos incurs severe memory and computational constraints. Network training and convergence is also highly unstable. We address these issue with following contributions:
\begin{itemize}
	\item Progressive growing of video generative adversarial networks
	\item Improved loss function for better stability
	\item Novel $300\times 300$ facial dynamics dataset with $10,910$ video clips
\end{itemize}

\section{Thesis Organization}
In chapter 2, readers will be introduced to basic ideas behind GANs and recent advances for stable training of GANs. In chapter 3, relevant literature will be reviewed and important models will be discussed in details. In chapter 4, proposed techniques for spatio-temporal growing of gans will be discussed. Sliced Wasserstein GAN loss for stable training will be presented in chapter 5. In chapter 6, various metrices used in this work for evaluating and comparing our model against other models will be presented. In chapter 7, we will discuss about novel dataset collected for this work. Additional datasets used for evaluating our model will be also be mentioned. Qualitative and quantitative comparision of our model with existing models will be presented in chapter 8. In chapter 9, we will discuss our findings and conclusions.
\chapter{Background}
For limited number of computer vision problems such as object recognition, collection of labelled data is relatively easy. Supervised machine learning techniques have already reached super human performance on such tasks. However, for several other problems such as segmentation, collection of labelled data is not as easy and the performance of purely supervised techniques is still at sub-human level. Semi-supervised and unsupervised techniques are promising avenue for such problems with information bottleneck. There have been numerous works in the direction of unsupervised representation learning. In particular, in computer vision community, various techniques such as Variational Autoencoders (VAE), Autoregressive models and Generative Adversarial Networks (GAN) have been utilized before for unsupervised image generation. Out of these techniques, GANs in particular have been main focus of the community for last couple of years. Though GANs suffered from stability of training and failure to converge in early years, these issues have largely been addressed with improved loss function \cite{wgan}\cite{iwgan}\cite{swgan}\cite{gmswd}, robust network architecture \cite{pggan}\cite{stackgan} and improved training algorithms \cite{ttur}\cite{wgan}.

In following sections, we will briefly introduce GANs and discuss different techniques proposed so far for stable training of GANs.

\section{GAN}
\begin{figure}
	\centering
	\includegraphics[width=.9\textwidth]{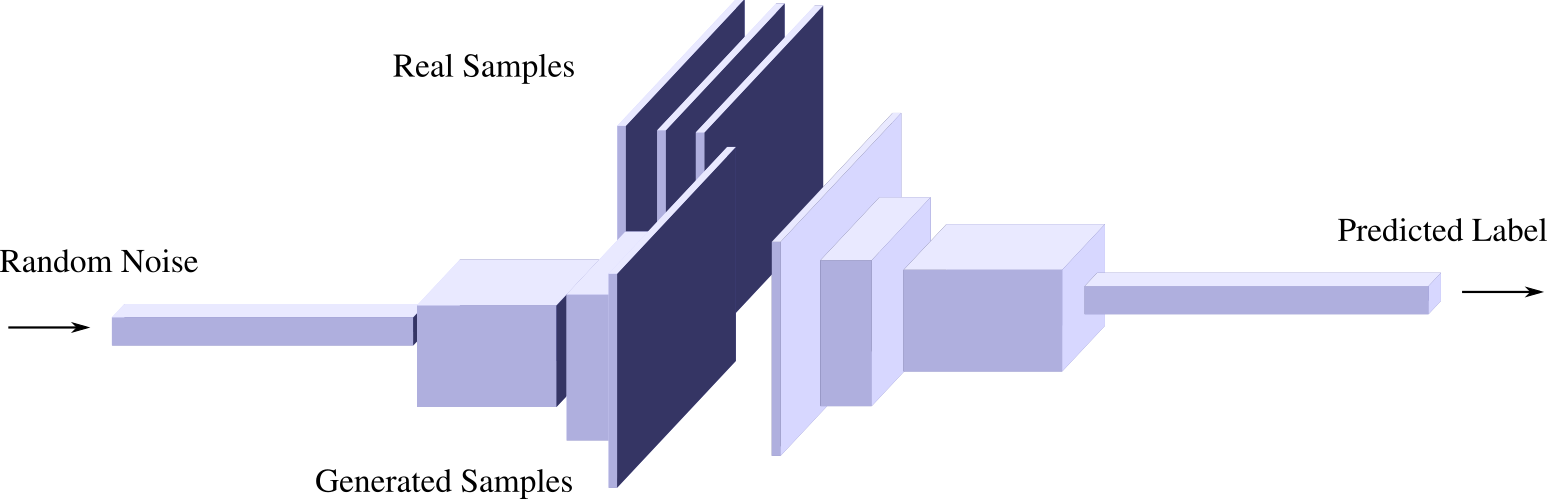}
	\label{fig:gan}
	\caption{Standard GAN Architecture.}
\end{figure}
Generative Adversarial Networks (GANs)\cite{ganivan} are unsupervised generative models that learn to generate samples from a given distribution in adversarial manner. The network architecture consists of generator $\G$ and discriminator $\D$ (in some cases also called critic). Given a random noise as input, the generator $\G:\R^k\rightarrow \R^m$, where $k$ is latent space dimension, tries to generate samples from a given distribution. The discriminator $\D$ tries to distinguish whether the generated sample is from a given distribution or not. The loss function is designed so that generator's goal is to generate samples that fool the discriminator. Similarly, the discriminator's goal is designed to avoid being fooled by the generator. As such, GANs can be interpreted as non-cooperative games. Let $\mathcal{G}:\z\in\R^k\rightarrow \x\in R^m$ be generator and $\mathcal{D}:\x\in\R^m\rightarrow \{0,1\}$ be discriminator. Then the loss function proposed in \cite{ganivan} is given by:
\begin{equation}
\mathcal{F}(\D,\G)=\E_{\x\sim p_\x}[-\log \mathcal{D}(\x)]+\E_{\z\sim p_\z}[-\log(1-\D(\G(\z)))],
\label{eq_ganloss}
\end{equation}
where $\z$ is latent code, $\x$ is data sample, $p_\z$ is probability distribution over latent space and $p_\x$ is probability distribution over data samples.
The two-player minimax game is then given by:
\begin{equation}
\min_{G}\max_{D} \mathcal{F}(\D,\G).
\end{equation}

In early years, discriminators were trained with sigmoid cross entropy loss as they were trained as classifiers for real and generated data. However, it was argued in \cite{lsgan}, that such GANs suffer from vanishing gradient problem. Instead, in \cite{lsgan}, least squares GANs were proposed that used least squares loss to train discriminators instead of sigmoid cross entropy loss.

Despite some improvements from least squares loss\cite{lsgan}, GANs still suffered from several issues such as instability in training, mode collapse and lack of convergence. In \cite{itgan}, authors proposed various techniques such as feature matching and minibatch discrimination. In feature matching, activations of intermediate layers of discriminator are used to guide the generator. Formally, the new generator loss is given by:
\begin{equation}
||\E_{\x\sim p_\x}\f(\x)-\E_{\z\sim p_\z}\f(\G(\z))||_2^2,
\end{equation}
where $\f$ replaces the traditional $\D$ in the form of a feature extractor rather than a discriminator. The discriminator is trained as usual. Minibatch discrimination technique was proposed to address the issue of mode collapse. To accomplish this, additional statistics that models affinity of samples in a minibatch is concatenated to features in the discriminator. it is important to note that the summary statistics is learned during training through large projection matrices.

\section{Wasserstein GAN}
In \cite{wgan}, authors argue that optimization of GAN loss given in Eq.~\ref{eq_ganloss} is same as minimization of Jensen-Shannon (JS) divergence between distribution of generated and real samples. In case the two distributions have non-overlapping support, JS-divergence can have jumps. This leads to aforementioned optimization issues. For stable training of GANs, authors propose to minimize Wasserstein Distance (WD) between  the distributions which behaves smoothly even in case of non-overlapping support. Formally, the primal form of WD is given by:
\begin{equation}
    W(p_r,p_g)=\inf_{\gamma\in\prod(p_r,p_g)}\E_{(x,y)\sim\gamma}[||x-y||],
    \label{wdprimal}
\end{equation}
where $p_r, p_g$ are distributions of real and generated samples and $\prod(p_r,p_g)$ is the space of all possible joint probability distributions of $p_r$ and $p_g$. It is not feasible to explore all possible values of $\gamma\in\prod(p_r,p_g)$. So, authors propose to use the dual formulation which is better suited for approximation. Formally, the dual formulation of Eq.~\ref{wdprimal} is given by:
\begin{equation}
    W(p_r,p_g) = \frac{1}{K}\sup_{||f||_L\leq K}\E_{\x\sim p_r}[\f(x)]-\E_{\x\sim p_g}[\f(\x)],
\end{equation}
where $||f||_L\leq K$ is $K-$Lipschitz constraint. The GAN loss is then given by:
\begin{equation}
    \mathcal{F}(\G, \D) = W(p_r,p_g)=\max_{w\in W}\E_{\x\sim p_r}[\f_w(\x)] - \E_{\z\in p_r(\z)}[\f_w(\G(\z))].
\label{eq_wganloss}
\end{equation}
Here, the discriminator takes the form of feature extractor and is parametrized by $w$. $\f$ is further required to be K-Lipschitz. In order to enforce the K-Lipschitz constraint, authors proposed weight clipping. However, as argued in the same paper, gradient clipping is a very rudimentary technique to enforce the Lipschitz constraint. In \cite{iwgan}, authors propose to penalize the norm of the gradient in order to enforce Lipschitz constraint. In particular, the new loss is defined as:
\begin{equation}
\mathcal{F}(\D,\G)=\E_{\x\sim p_\x}[\D(x)]-\E_{\z\sim p_\z}[\D(\G(\z))] + \lambda \E_{\hat{\x}\sim p_{\hat{\x}}}[(||\nabla_{\hat{\x}}\D(\hat{\x})||_2-1)^2]
\label{eq_iwganloss},
\end{equation}
where $\lambda$ is regularization parameter and $\hat{\x}$ is sampled uniformly from a straight line connecting real sample and generated sample.

\section{Conditional GANs}
It is relevant to briefly review Conditional GANs\cite{conditiongan}. They provide a framework to enforce conditions on the generator in order to generate samples with desired property. Such conditions could be class labels or some portion of original data as in case of future prediction. Under the new setting, the GAN loss defined in Eq.~\ref{eq_ganloss} becomes
\begin{equation}
\mathcal{F}(\D,\G)=\E_{\x\sim p_\x}[-\log \mathcal{D}(\x)]+\E_{\z\sim p_\z}[-\log(1-\D(\G(\z)))]
\label{eq:condtganloss}
\end{equation}
Concrete applications of Conditional GANs include generation of specific digits of MNIST dataset \cite{conditiongan} or a face with specific facial expression, age or complexion in the context of images \cite{stargan}, or future prediction\cite{ivgan} in context of videos to name a few.

Over the years, GANs have found applications in numerous areas. To name a few, such applications include image-to-image translation\cite{cyclegan}\cite{cyclecondt}\cite{bicyclegan}\cite{stargan}, 3D object generation\cite{3dgan}\cite{3dgan2}, super resolution\cite{superresolution1} , image inpainting\cite{inpaint1} etc.
%
\newpage
\chapter{Related Works}
As outlined in the title, our focus on this work is on higher resolution video generation with progressive growing of Sliced Wasserstein GANs (SWGAN). As such, we will briefly discuss existing works related to progressive growing technique, video generation and Sliced Wasserstein GANs in the following sections. Later, we will also discuss the details of the selected works relevant to this work.

As the complexity of the problem grows, it becomes more and more difficult to learn an appropriate model. To address this, the idea of curriculum learning was proposed in~\cite{curriculum}. The idea behind curriculum learning is to gradually increase the complexity of the problem during training. The idea to use multiple generators has been explored in~\cite{multigan} to address the issue of mode collapse. In~\cite{multidisc}, authors proposed to use multiple discriminators for stable training. Similarly, ~\cite{stackgan} introduces multi-stage GANs where consecutive GANs take input from GANs from previous stage. The details and complexity of features of the generated samples increases with increasing stages of GANs. This was one of the  early GANs to generate reasonable images of size 256x256. Inspired by these concepts of using multiple generators and discriminators, and curriculum learning, authors of \cite{pggan} proposed the technique of progressively growing the network for stable generation of 1024x1024 images.

There have been numerous works that aim to improve the original objective function for GANs Eq.~\ref{eq_ganloss}. This has already been reviewed in detail in earlier chapter. LS-GAN\cite{lsgan}, WGAN\cite{wgan}, I-WGAN\cite{iwgan} are one of the most important works that aim to improve the original GAN loss function. Recently, Sliced Wasserstein GANs (SWGANs) that propose to use Sliced Wasserstein Distance (SWD) were proposed in \cite{swgan}\cite{gmswd}. As stated in~\cite{bonnette}, SWD is equivalent to WD. However, SWD is relatively easier to approximate compared to original WD\cite{kolouri}\cite{bonnette}. SWGANs are motivated by the fact that WD is ideal to measure distance between two distributions with non-overlapping support and enforcing the Lipschitz constraint on the dual formulation in non trivial.

In the direction of video generation, there are very limited works that try to model videos in unsupervised setting. One of the earliest works~\cite{vgan} tried to separate background and foreground for video generation. In~\cite{vpn}, Video Pixel Network (VPN) was proposed building on the work of Pixel CNNs. Temporal GAN~\cite{tgan} used temporal and image generators to generate temporal encoding and images separately. MoCoGAN~\cite{mocogan} also use temporal and spatial generators. However, they claim to separate motion and content of video using different techniques to encode latent space. One recent work~\cite{ftgan} tries to model videos by separating texture and optical flow. All of these works generate videos of 64x64 resolution. In this work we generate videos of up to 256x256 resolution. It is important to note that most of the above mentioned works design their model to do some sort of separation on the space of videos with different names such as foreground and background separation or texture and flow separation or motion and content separation. In this work, we do not carry any such separation entirely for model simplicity in light of model complexity introduced by progressively growing scheme. However, advantage of single stream models while generating unstabilized videos has been highlighted in \cite{ivgan}.

As the ideas behind progressive growing techniques\cite{pggan} and video generative networks\cite{tgan}\cite{vgan} are relevant to this work, these works will be discussed in detail in following section.
\section{Progressive Growing of GAN}
As mentioned earlier, the basic idea behind progressive growing of GANs is to gradually increase the complexity of the problem \cite{pggan}. To accomplish this, authors propose to first train the generators and discriminators on lower resolution samples. During training they propose to progressively introduce new layers to increasingly learn on more complex problem of generating higher resolution images.
\begin{figure}
	\centering
	\includegraphics[width=.65\textwidth]{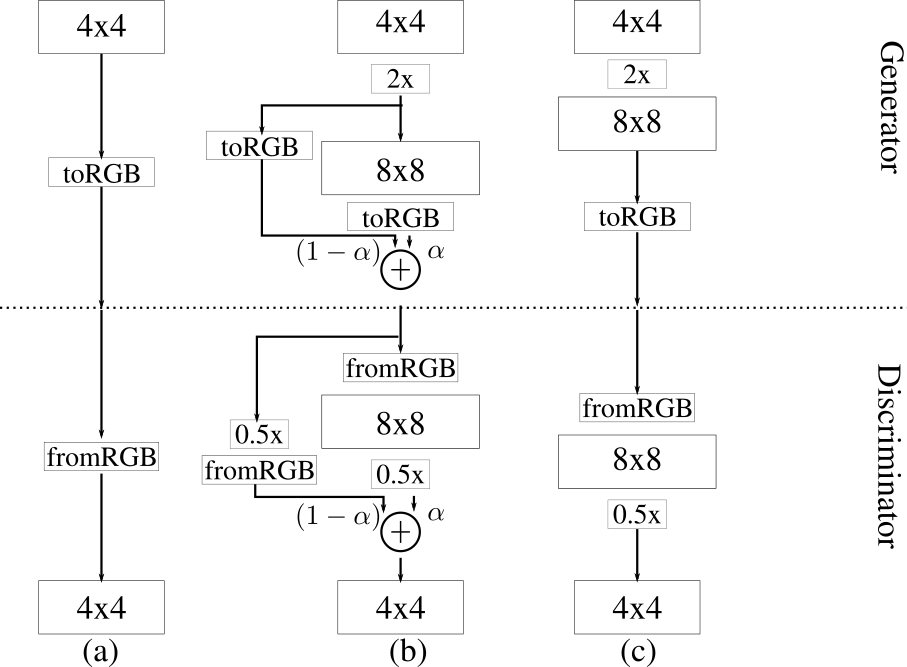}
	\caption{Transition phase during growing of generator and discriminator networks under the progressive growing scheme.}
	\label{fig:progan}
\end{figure}
As illustrated in Fig.~\ref{fig:progan}, successively new networks layers are introduced to generators and discriminators. The transition from one stage of training to another stage of training is made smooth by using linear interpolation during transition phase. The interpolation factor is then smoothly changed during training. At every stage, the output layer of the generator consists of a 1x1 convolutions that map feature channels to RGB image. Similarly, the first layer of discriminator consists of 1x1 convolutions that map RGB image to feature channels. During the transition step, a linear interpolation of output of 1x1 convolutions from lower resolution feature channels and 1x1 convolutions from higher resolution feature channels is taken as output of generator. The scalar factor $\alpha$ corresponding to output of higher resolution feature channels is smoothly increased from $0$ to $1$. Similarly, during transition, both higher resolution and downscaled images are provided as input to different input layers of discriminator. Learning on simpler problem and gradually increasing complexity of the problem for both discriminator and generator can be expected to lead to faster convergence and stability. Authors claim in the paper that the improvement with proposed training scheme is orthogonal to improvements arising from loss functions. The idea of progressive growing has not yet been applied to video generation. In this work, we explore progressively growing of video generative networks.

\section{Video GAN}
\begin{figure}[!ht]
	\centering
	\includegraphics[width=.75\textwidth]{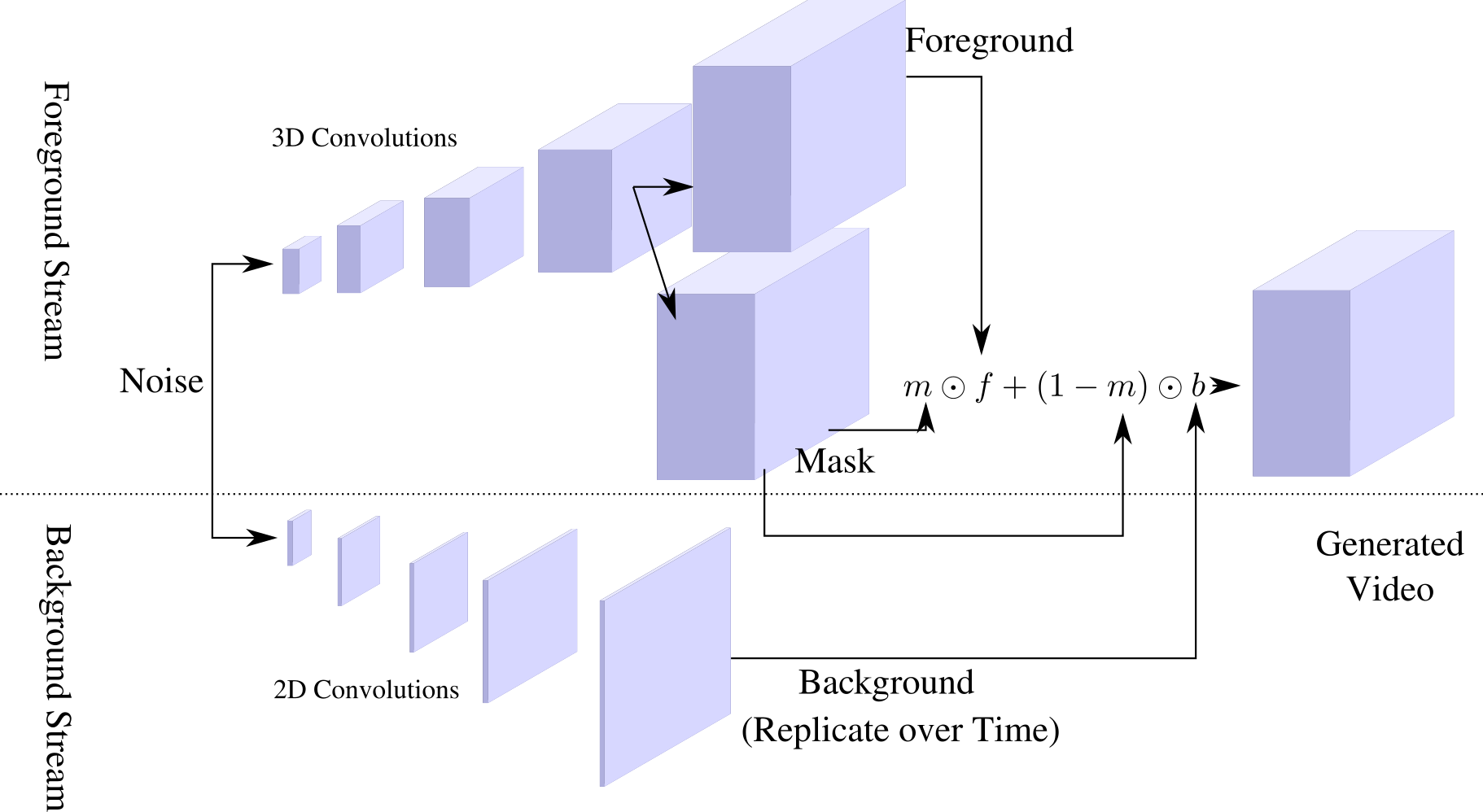}
	\caption{Two stream architecture of the generator of Video GAN. Video GAN assumes stable background in the video clips.}
	\label{fig:videogan}	
\end{figure}

In \cite{vgan}, authors propose a parallel architecture for unsupervised video generation. The architecture consists of two parallel streams consisting of 2D and 3D convolution layers for the generator and single stream 3D convolution layers for discriminator. As illustrated in Figure~\ref{fig:videogan}, the two stream architecture was designed to untangle foreground and background in videos.
If $0\leq m(z)\leq 1$ be the mask that selectes either foreground or background, the output of generator, $\G$, at pixel $z$ is given by:
\begin{equation}
    \G(z)=m(z)\odot f(z) + (1-m(z))\odot b(z),
\end{equation}
where $b(z)$ is output of background stream and $f(z)$ is output of foreground stream. In case of background stream, the same value of $b(z)$ is replicated over all time frames. Experimental results presented by authors supports the use of two stream architecture. However, one of the strong assumptions of the model is that of static background.
\section{Temporal GAN}
\begin{figure}[!ht]
	\centering
	\includegraphics[width=.75\textwidth]{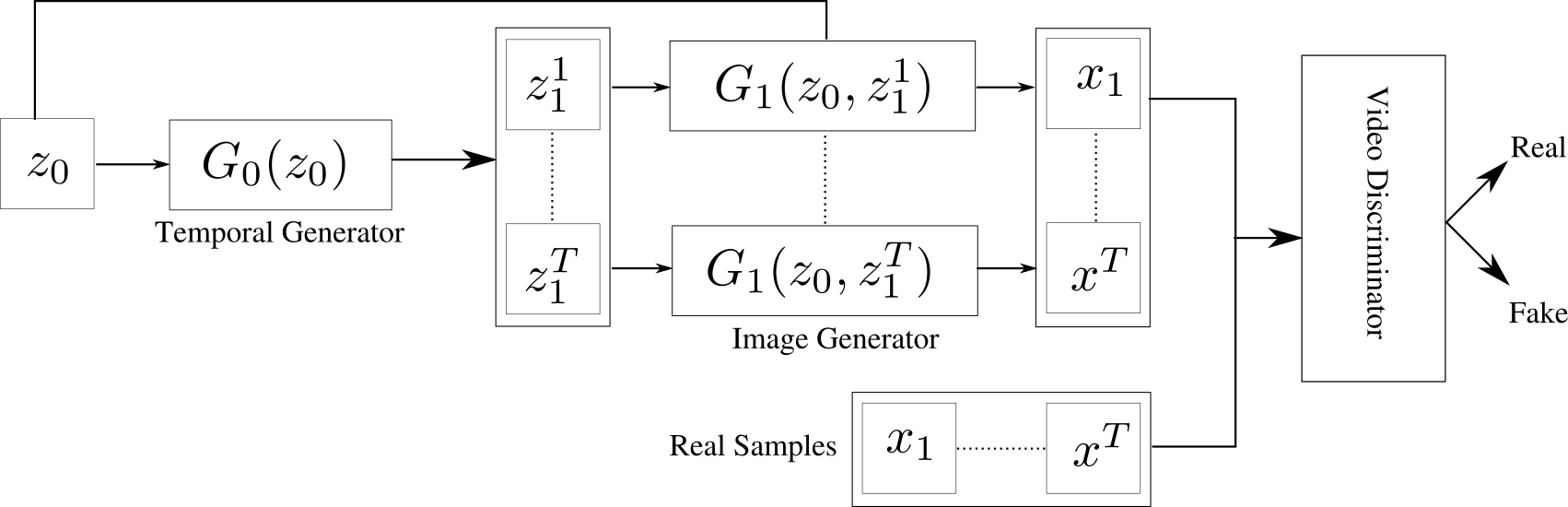}
	\caption{Temporal GAN uses a cascade architecture to generate videos. Temporal generator uses 1-D deconvolutions and spatial generator uses 2-D deconvolutions.}
	\label{fig:tgan}	
\end{figure}

In \cite{tgan}, authors propose a cascade architecture for unsupervised video generation. As illustrated in Fig.~\ref{fig:tgan}, the proposed architecture consisting of temporal and image generator. Temporal generator, which consists of 1-D deconvolution layers, maps input latent code to a set of new latent codes corresponding to frames in the video. Each new latent code and the original latent code together are then fed to a new image generator. The resulting frames are then concatenated together to obtain a video. For the discriminator, TGAN uses single stream 3D convolution layers.

Unlike in the case of Video GAN, this model makes no assumption about separation of background and foreground stream. As such, no requirement on background stabilization of videos is assumed.
\section{MoCoGAN}
Motion Content GAN (MoCoGAN) network architecture is similar to TemporalGAN (TGAN) \cite{tgan} in the sense it also has cascade architecture. Furthermore, it also uses temporal and image generators and 3D convolution layers based discriminator. However, unlike TGAN, temporal generator on MoCoGAN is based on Recurrent Neural Network (RNN) and the input to temporal generator is a set of latent variables. Furthermore, the outputs of temporal generator, motion codes, are concatenated with newly sampled content code to feed image generators. In discriminator, there is an additional image based discriminator. The authors claim that using such architecture helps to separate motion and content from videos.
\section{Other Related Works}
Video Pixel Networks (VPN) \cite{vpn} build on the work of PixelCNNs \cite{pixelcnn} for future prediction. In particular, they estimate probability distribution of raw pixel values in video using resolution preserving CNN encoders and PixelCNN decoders. Other works for future prediction include \cite{lecun}. Recently, optical flow based models have produced more realistic results. In particular, in \cite{ftgan} authors use flow and texture GAN to model optical flow and texture in videos. Several of the works have also focused on future prediction which is a slightly different problem than unsupervised video generation\cite{ivgan}\cite{lecun}.
\chapter{Progressive Video Generation}
GANs typically suffer from instability and failure to converge. Such issues are even more prominent for higher resolution images or video generation as the generator and discriminator contain too many parameters. In such cases improved loss function itself may not suffice to generate high resolution videos.
\begin{figure}[!ht]
	\centering
	\includegraphics[width=.8\textwidth]{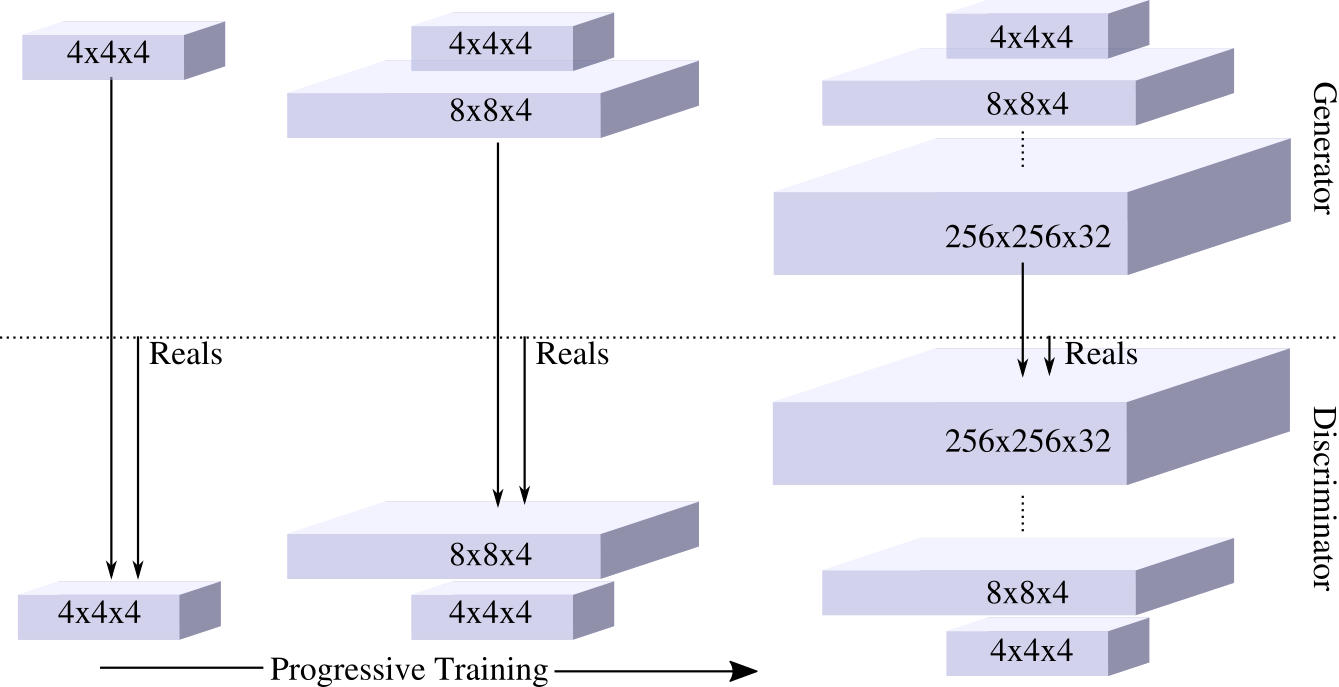}
	\label{fig:provid2}
	\caption{Progressive video generation. Initially low resolution and short videos are generated. Gradually, higher resolution and longer videos are generated.}
\end{figure}

To address this issue, the idea behind curriculum learning \cite{curriculum} can be utilized. In the beginning, smaller network with less parameters can be used to learn the lower resolution samples \cite{pggan}. Generator and discriminator can be trained to generate and discriminate downscaled videos. Learning a coarser model is relatively easier and turns out to be more stable. First learning simpler models and gradually increasing model complexity also leads to faster convergence. To increase model complexity during training, gradually new layers can be introduced both to the generator and the discriminator to generate larger resolution videos. Doing so helps the model to learn a finer distribution of samples. Progressively growing the network during training, helps to first estimate a coarser PDF and gradually refine it during training.

In order to progressively grow the network for video generation, the real $32\times256\times256$ video samples are downscaled to $4\times 4\times 4$ by applying 3-D average pooling filter. The generator then generates $4\times 4\times 4$ videos. New layers added during training gradually introduce more spatial and temporal details.

\section{Transition Phase}
\begin{figure}[!ht]
	\centering
	\includegraphics[width=.8\textwidth]{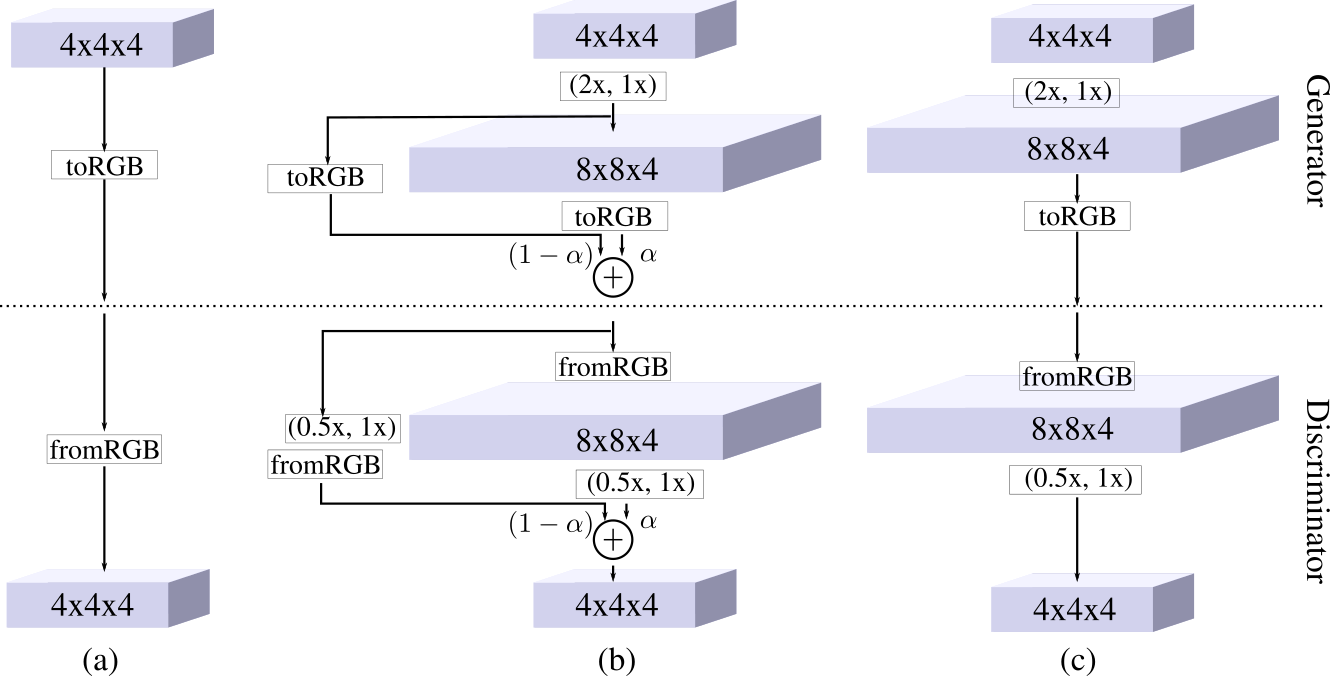}
	\label{fig:provid1}
	\caption{Transition phase during which new layers are introduced to both generator and discriminator.}
\end{figure}

During each phase, the final layer of generator consists of $1\times 1 \times 1$ convolution filters that map input feature channels to RGB videos. The discriminator in the similar fashion consists of $1\times 1 \times 1$ convolution filters that map input RGB videos to feature channels. While transitioning from one resolution to another resolution, new convolution layers are introduced to both discriminator and generator symmetrically to generate larger resolution videos. During transition from one level of detail to another level of detail, generator outputs videos of two different resolutions. The lower resolution videos are upscaled with nearest-neighbor upscaling. The linear combination of the upscaled video and higher resolution video is then fed to discriminator. The weight corresponding to higher resolution video generated by generator is smoothly increased from 0 to 1 and that corresponding to upscaled video is gradually decreased from 1 to 0. New layers are introduced in discriminator in similar manner.

\section{Minibatch Standard Deviation}
One way to avoid mode collapse is to use feature statistics of different samples within the minibatch and penalize the closeness of those features \cite{iwgan}. In this approach, the feature statistics are learned through parameters of projection matrices that summarize input activations \cite{pggan}\cite{wgan}. Instead, following \cite{pggan}, standard deviation of individual features from each spatio-temporal location across the minibatch is computed and then averaged. Thus obtained single summary statistics is concatenated to all spatio-temporal location and features of the minibatch.
\begin{figure}[!ht]
	\centering
	\includegraphics[width=.6\textwidth]{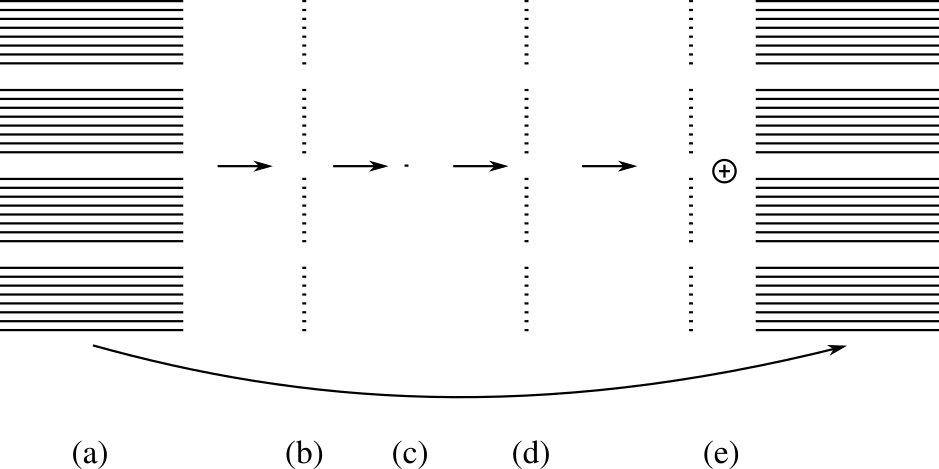}
	\label{fig:provid1}
	\caption{Illustration of different steps of minibatch standard deviation layer: (a) feature vectors at each pixel across the minibatch, (b) standard deviation computation of each feature vector, (c) average operation, (d) replication and (e) concatenation.}
\end{figure}

Since there are no additional learnable parameters, this approach is computationally cheaper and yet as argued in \cite{pggan}, efficient.

\section{Pixel Normalization}
Following \cite{pggan} and in the direction of local response normalization proposed in \cite{alexnet}, normalization of feature vector at each pixel avoids explosion of parameters in generator and discriminator. The pixel feature vector normalization proposed in \cite{pggan} can be naturally extended to spatio-temporal case. In particular, if $a_{x,y,t}$ and $b_{x,y,t}$ be original and normalized feature vector at pixel $(x,y,t)$ corresponding to spatial and temporal position,
\begin{equation}
    b_{x,y,t}=\frac{a_{x,y,t}}{\sqrt{\frac{1}{N}\sum_{j=0}^{N-1}(a^j_{x,y,t})^2+\epsilon}}, 
\end{equation}
where $\epsilon=10^{-8}$ and $N$ is number of feature maps. Though pixel vector normalization may not necessarily improve performance, it does avoid explosion of parameters in the network.
\chapter{Sliced Wasserstein GAN Loss}
As discussed earlier chapters, in \cite{wgan}, authors proposed to
approximate Wasserstein Distance (WD) using it's dual formulation. The dual formulation was obtained using Kantorovich-Rubinstein duality. Thus obtained formulation has form of a saddle-point problem and is usually difficult to optimize \cite{deshpande}. Instead of the above discussed formulation, Sliced Wasserstein Distance (SWD) can be used to measure the distance between two distributions. It was proven in \cite{bonnette} that SWD is equivalent to WD. To compute the SWD, the plane is sliced using lines passing through the origin and higher dimensional marginal distributions are projected onto these lines using Radon transform. The Radon transform is performed using orthogonal projection matrices. 
\begin{figure}[!ht]
	\centering
	\includegraphics[width=.8\textwidth]{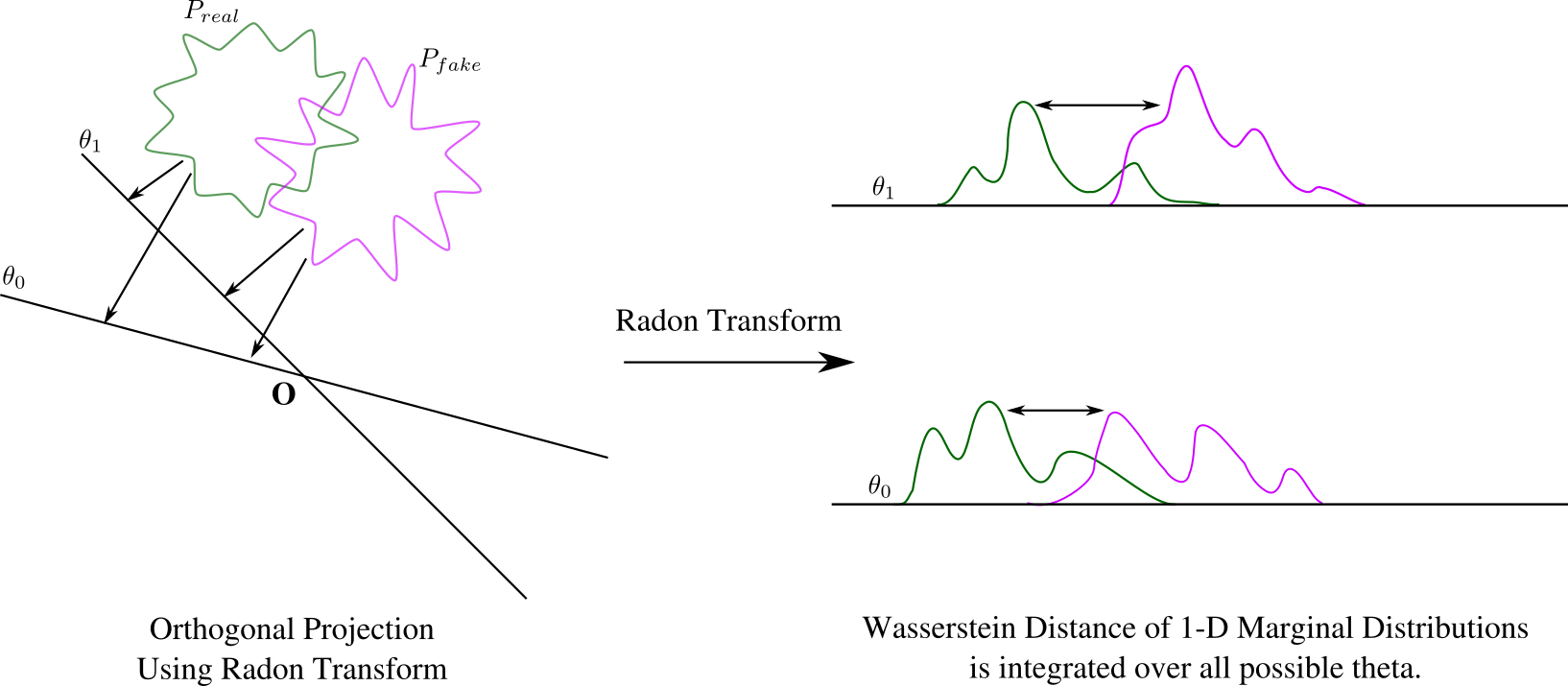}
	\label{fig:swgan}
	\caption{Intuition behind Sliced Wasserstein Distance. After projection, the Wasserstein Distance of 1-D marginals is integrated over all possible values of $\theta$.}
\end{figure}
The required metric is then given by integral of the projections along all such lines. As the projected marginal distributions are one dimensional, the SWD takes the form of functional of 1-D WD. Since closed form solution of 1-D WD exists, approximating SWD becomes easier. Mathematically, the SWD as a functional of 1-D WD is given by
\begin{equation}
\int_{\mathbb{S}^{N-1}}
\left(\sup_{f \in \mathrm{L}^1} \mathbb{E}_{\bm{X}_{\theta} \sim P_{X_{\theta}}} [f(\bm{X}_{\theta})] - \mathbb{E}_{\bm{Y}_{\theta} \sim P_{Y_{\theta}}}  [f(\bm{Y}_{\theta})]\right)
d \theta,
\label{eq:swd}    
\end{equation}
where $L^1$ is the function space of all $1-$Lipschitz functions, $P_{X_\theta},P_{Y_\theta}$ are the projected marginal distributions discussed earlier. As the latent space is usually low-dimensional in GANs, it is implicitly assumed that the distribution of real samples lies on low-dimensional manifold. Hence, if $x=[x_1, x_2, \dots, x_n]\in\R^{N\times n}$ be $N$ dimensional $n$ input samples, using standard discriminator setting, input data is encoded to $K$ dimensional latent code $y = [y_1,y_2,\dots,y_n]\in \R^{K\times n} $. Then orthogonal transform matrices $\theta=[\theta_1,\theta_2,\dots,\theta_K]\in\R^{K\times K}$ are applied to project the $K-$dimensional encodings into $K$ one dimensional marginal distributions. The $k-$Lipschitz mapping function $f$ is given by:
\begin{equation}
f(\bm{y}) =
\begin{pmatrix}  
\phi(\lambda_{1} (\bm{\theta}_1^T \bm{y}) + b_{1}) \\
\vdots \\
\phi(\lambda_{K} (\bm{\theta}_K^T \bm{y}) + b_{K})
\end{pmatrix},
\label{eq:orth}
\end{equation}
where $\theta_i$ are projection matrices defined earlier. $\phi$ is an activation function, $\lambda_i$ and $b_i$ are scalars.
In practice, we compute $f(\bm{y}) = \frac{1}{K}\sum_{i=1}^{K}(\phi(\lambda_{i} (\bm{\theta}_1^T \bm{y}) + b_{i}))$ to approximate the integral of Eq.~\ref{eq:swd}, and the mapping function of the discriminator is $D=f \circ E$.
To avoid gradient explosion and vanishing for $E$, we additionally imposing the gradient regularizer on it. The final objective function is given by:
\begin{equation}
\begin{aligned}
\min_{\G} \max_{D} & \, \mathbb{E}_{\bm{X} \sim \mathbb{P}_X} [D(\bm{X})]  -\mathbb{E}_{{\bm{Z}} \sim \mathbb{P}_Z} [D(G(\bm{Z}))]
\\ & + \lambda_1 \mathbb{E}_{\bm{\hat{X}}
	\sim P_{\hat{X}}}
[\|\nabla_{\bm{\hat{X}}} E(\bm{\hat{X}})\|_2^2]
\\ & + \lambda_2 \mathbb{E}_{\bm{\hat{Y}}
	\sim P_{\hat{Y}}}
[(\|\nabla_{\bm{\hat{Y}}} f(\bm{\hat{Y}}))\|_2-k)^2],
\end{aligned}
\label{eq:swgan}
\end{equation}
where we sample the $\bm{\hat{X}}, \bm{\hat{Y}}$ based on \cite{iwgan}, where $\lambda_1, \lambda_2$ are the coefficients to balance the penalty terms. $\lambda_2$ is also used to absorb the scale $k$ caused by the $k$-Lipschitz constraint.
\chapter{Evaluation Metrices}
Evaluation of GANs is a non trivial problem. Some of the early works relied on evaluation based on surveys such as Amazon Mechanical Turk~\cite{itgan}\cite{vgan}\cite{ivgan}. More quantitative metrics such as Inception Score~\cite{itgan} and Frechet Inception Distance (FID) \cite{ttur} have been proposed for image based GAN evaluation. These metrices were shown to correlate well with human perception. We will briefly review these metrices in following sections.
\section{Inception Score}
Inception score was originally proposed in~\cite{itgan} for evaluation of GANs. In the paper, the authors argued that Inception Score correlated well with the visual quality of generated samples. Let $\mathbf{x}\sim \mathcal{G}$ be samples generated by the generator $\mathcal{G}$. $p(y|\mathbf{x})$ be the distribution of classes for generated samples and $p(y)$ be the marginal class distribution:
\begin{equation}
p(y)=\int_{\mathbf{x}}p(y|\mathbf{x})p_g(\mathbf{x}).
\end{equation}
The Inception score is defined as:
\begin{equation}
IS(\G)=\exp(\mathbb{E}_{\mathbf{x}\sim p_g}\D_{KL} (p(y|\mathbf{x})||p(y))),
\end{equation}
where $\D_{KL}$ is the Kullback-Leibler divergence between $p(y|\mathbf{x})$ and $p(y)$.

In practice, the marginal class distribution is approximated with:
\begin{equation}
\hat{p}(y)=\frac{1}{N}\sum_{i=1}^Np(y|\x^{(i)}),
\end{equation}
where $N$ is number of samples generated.

Intuitively, maximum Inception Score is obtained when generated samples can be clearly classified as belonging to one of the classes in training set and the distribution of samples belonging to different classes is as uniform as possible. This encourages realistic samples and discourages mode collapse. The idea behind Inception score has been generalized to the context of video generation. In \cite{tgan} authors propose to use C3D model trained on Sports-1M dataset and finetuned on UCF101 dataset. It is important to point out that Inception score computation requires a model trained on specific classification problem and corresponding data. Furthermore Inception score does not compare the statistics of generated samples directly with statistics of real samples \cite{ttur}.
\section{Fr{\'e}chet Inception Distance}
Alternative measure to access the quality of generated samples was proposed in~\cite{ttur}. In the paper, authors propose to use pre-trained networks as feature extractors to extract low level features from both real and generated samples. If $\D$ be the CNN used to extract features, $(m_r, \Sigma_r)$ be mean and covariance of features extracted from real samples and $(m_f,\Sigma_f)$ be mean and covariance of features extracted from fake samples with $\D$, then the Fr{\'e}chet distance is defined as 
\begin{equation}
d^2((m_r,\Sigma_r),(m_f,\Sigma_f))=||m_r-m_f||_2^2+Tr(\Sigma_r+\Sigma_f-2(\Sigma_r\Sigma_f)^{1/2}).
\end{equation}
FID was shown to correlate well to visual perception quality~\cite{ttur}. Since FID directly compares the summary statistics of generated samples and real samples, it can be considered to be more accurate than Inception score. Furthermore, as lower level features are used to compute FID score, it can be used to evaluate generative models for any dataset.

Similar to Inception score, FID can be generalized to compare video generative models. In particular, as C3D is standard model widely used in video recognition tasks, a C3D model trained on action recognition dataset can be used as feature extractor. Since output of final pooling layer is very high dimensional in case of C3D, output of first fully connected layer can be used to resonably compare the quality of generated samples.
\chapter{Evaluation Datasets}
To evaluate our video generative mode, we collected our own TrailerFaces dataset. Furthermore, we also evaluated our model on existing Golf~\cite{vgan}, Aeroplane~\cite{ivgan} and UCF101 datasets~\cite{ucf101_dataset}. In following sections, we will review these datasets briefly.
\section{Trailer Face Dataset}
\begin{figure}[!ht]
	\centering
	\includegraphics[width=.9\textwidth]{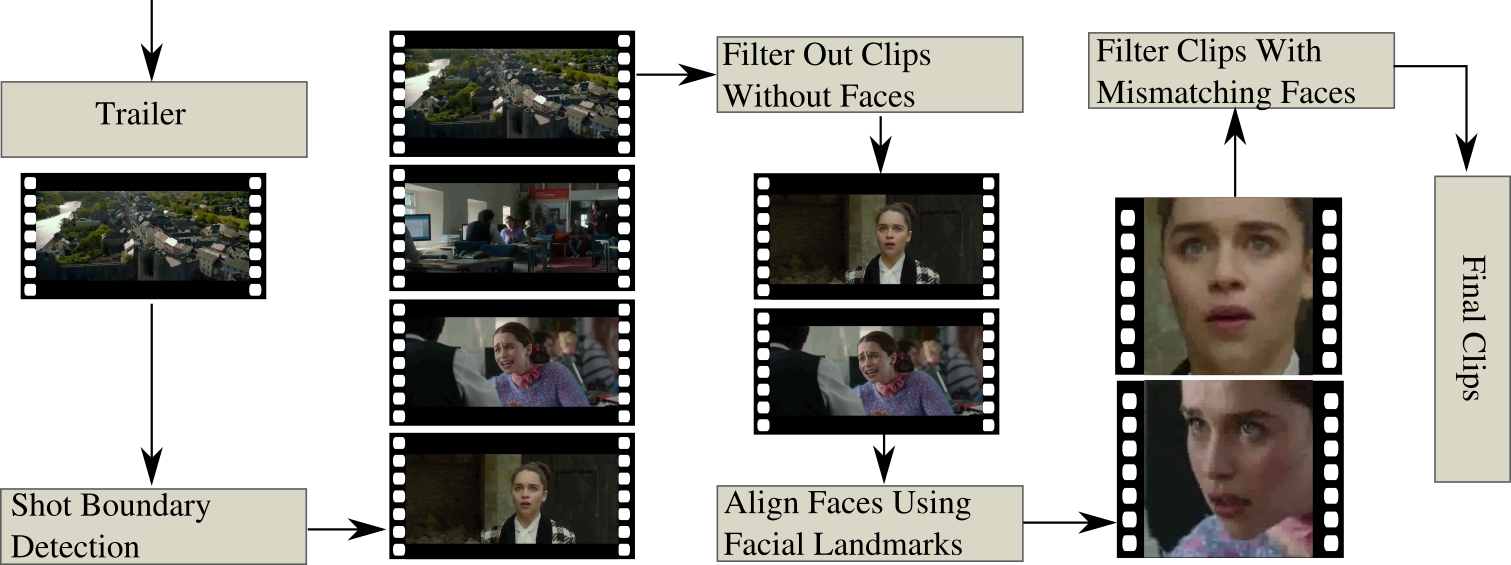}
	\label{fig:pipeline}
	\caption{Pipeline for Construction of Dataset.}
\end{figure}
Large chunk of GAN papers evaluate and compare the generative models on facial datasets \cite{pggan}\cite{ttur}\cite{ganivan} such as CelebA\cite{celeba_dataset} in case of images and MUG dataset\cite{mug_dataset} or YouTube Faces\cite{youtubefaces_dataset} in case of videos~\cite{mocogan}\cite{rehuang2017face}. However, there is lack of publicly available high resolution datasets containing facial dynamics.
\begin{table}
	\begin{center}
		\begin{tabular}{lccccc}
			\hline\noalign{\smallskip}
			Dataset  & Resolution (Aligned) & Sequences & Wild & Labels & Diverse Dynamics\\
			\noalign{\smallskip}
			\hline
			\noalign{\smallskip}
			TrailerFaces &  300x300  & 10,911 &\ding{51} &\ding{55}&\ding{51}\\
			YoutubeFaces &  100x100  & 3,425 & \ding{51} & Identity&\ding{55}\\
			AFEW & - & 1,426 & \ding{51} & Expressions & \ding{51} \\
			MUG & 896x896 & 1,462 & \ding{55} & Expressions & \ding{55}\\ \hline
		\end{tabular}
		\caption{Comparision of our TrailerFaces dataset with existing datasets containing facial dynamics.}
		\label{table:dataset_comparision}
	\end{center}
\end{table}
In terms of resolution too, widely used datasets for video generation such as Golf and Aeroplane datasets too are only 128x128 resolution. UCF101 is widely used for evaluation of generative models. Though it contains 240x320 resolution samples, due to relatively small number of samples per class, learning meaningful features is not possible. Aeroplane and Golf datasets contain too diverse videos. Learning meaningful representation from such videos can be difficult for networks. Hence a novel dataset of human facial dynamics was collected from movie trailers.
\begin{table}[!ht]
	\begin{center}
		\begin{tabular}{lcccccc}
			\hline\noalign{\smallskip}
			Number of Frames  & 30-33 & 34-39 & 40-47 & 48-57 & 58-69 & 70-423 \\
			\noalign{\smallskip}
			\hline
			\noalign{\smallskip}
			Total clips &  1781 & 3106 & 2291 & 1591 & 940 & 1201\\ \hline
		\end{tabular}
		\caption{Total number of clips with given number of frames.}
		\label{table:dataset_frequency}
	\end{center}
\end{table}

Our motivation to use movie trailers for dataset collection was motivated by the fact movie trailers highlight dramatic and emotionally charged scenes. Unlike whole movies, interviews or TV series, trailers contain scenes where stronger emotional response of actors are highlighted. Furthermore using trailers of thousands of movies increases the gender, racial and age-wise diversity of the faces in the clips. Approximately $6000$ complete Hollywood movie trailers were downloaded from YouTube. Number of SIFT feature matches between corresponding frames was used for shot boundary detection. Approximately $200,000$ shots were detected in those trailers. After splitting trailers into individual shots, those with too few or too many frames were removed. Face-detection was carried out to filter-out clips where at least 31 consecutive frames do not contain any faces. For face detection Haar-cascade based face detection tool from Open-CV was used. After detection of faces, Deep Alignment Network\cite{deepalignmentnetwork} was used for extraction of $68$-point facial landmark. Thus obtained facial landmarks were used for alignment using similarity transform. This was observed to be more stable across temporal dimension compared to state-of-art techniques like MTCNN. Finally, consecutive frames from those shots on which face detection was successful were selected. SIFT feature matching was again used to remove clips containing different personalities across frames.
\begin{figure}[!ht]
	\centering
	\includegraphics[width=.9\textwidth]{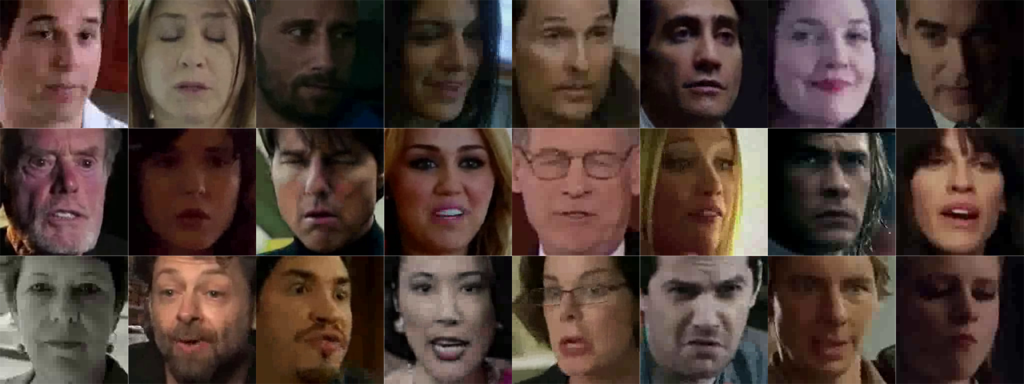}
	\label{fig:trailerFacesSamples}
	\caption{Samples from TrailerFaces Dataset. Random frames were chosen for visualization.}
\end{figure}

\section{UCF101 Dataset}
UCF101 dataset~\cite{ucf101_dataset} was originally collected for action recognition tasks. It contains 13320 videos from 101 different action categories. Some of the action categories in the videos include Sky Diving,  Knitting and Baseball Pitch. In \cite{tgan} video based inception score was proposed for evaluation of quality of video generative models. As argued in \cite{tgan}, Inception score computation requires a dataset with class labels and a standard model trained for classification. For training, first training split of UCF101 dataset with 9,537 video samples was used.
\begin{figure}[!ht]
	\centering
	\includegraphics[width=.9\textwidth]{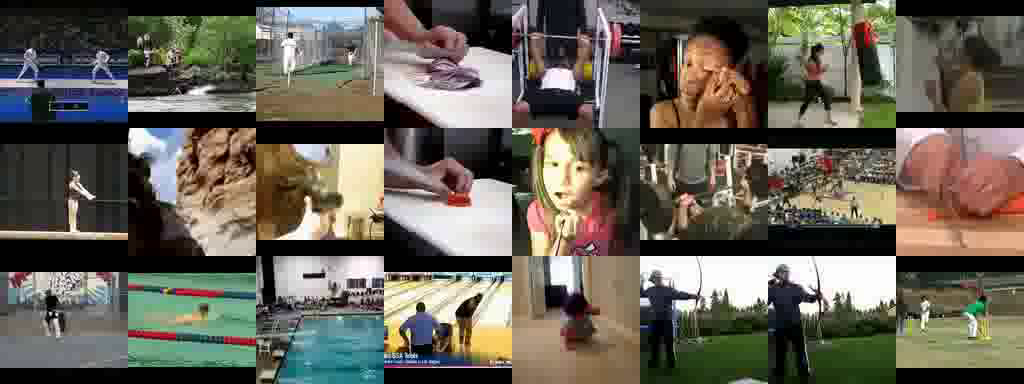}
	\caption{Samples from UCF101 dataset. Random frames were selected for visualization.}
	\label{fig:ucf101}	
\end{figure}

\section{Golf and Aeroplane Datasets}
Golf and Aeroplane datasets contain 128x128 resolution datasets that can be used for evaluating video generative adversarial networks. Golf dataset in particular was used in \cite{tgan}\cite{vgan}\cite{ivgan}. Both of these datasets contain videos in the wild. Golf dataset contains more than half a million clips. We used the background stabilized clips for training our model. Aeroplane dataset contains more than 300,000 clips that are not background stabilized.
\begin{figure}[!ht]
	\centering
	\begin{subfigure}
		\centering
		\includegraphics[width=0.44\textwidth]{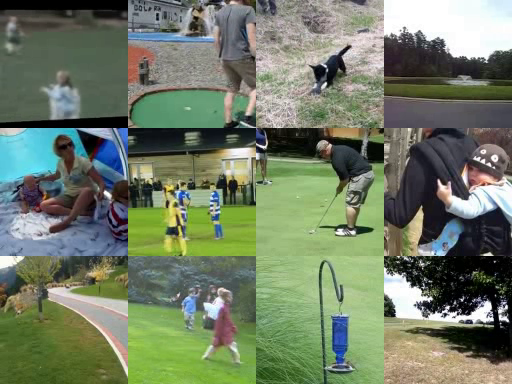}
	\end{subfigure}%
	\begin{subfigure}
		\centering
		\includegraphics[width=0.44\textwidth]{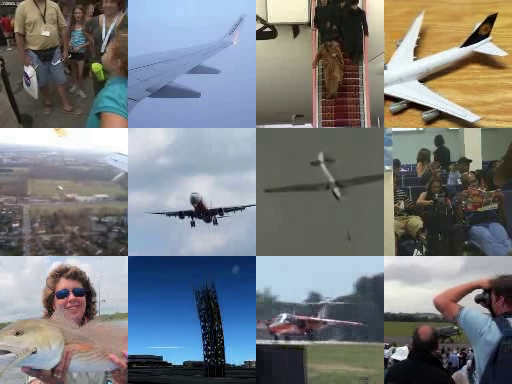}
	\end{subfigure}
	\label{fig:ucfgraphs}
	\caption{Samples from Golf (right) and Aeroplane (left) datasets. Random frame was selected for visualization.}
\end{figure}
%
\newpage
\chapter{Experiments and Results}
To compare the performance of our model with state of art models, we present qualitative and quantitative results below. Since UCF101 has class labels and as it was used for evaluation in prior works, we evaluate our models by comparing Inception score on UCF101 dataset with compare Golf and Aeroplane datasets with FID score. For TrailerFaces dataset, we present qualitative results below.

\section{Qualitative Results}
\begin{figure}[H]
	\centering
	\includegraphics[width=0.99\textwidth]{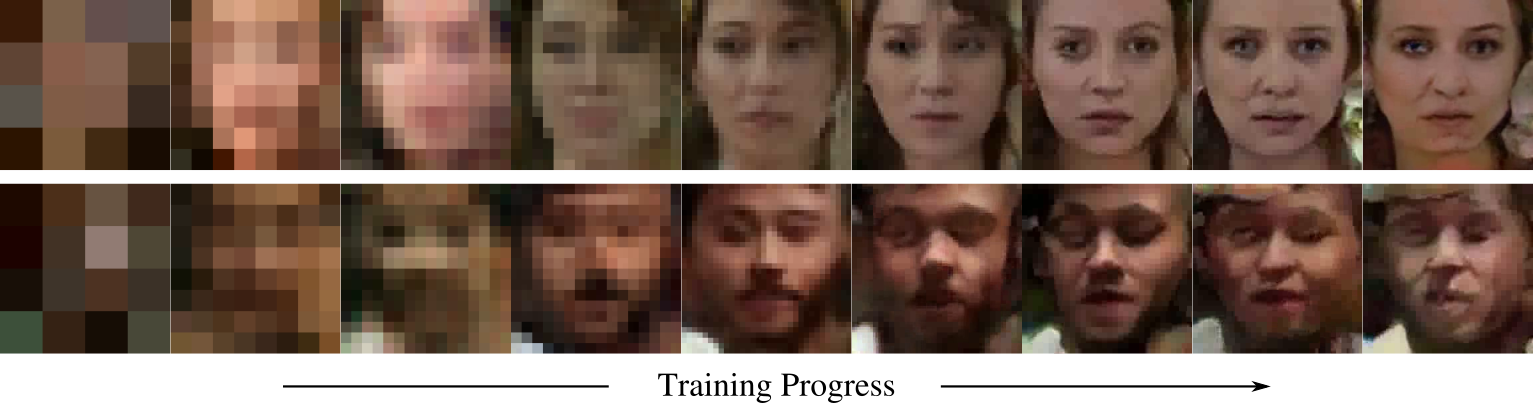}
	\caption{Improvement in resolution of generated videos over time on TrailerFaces dataset. Single frames from each video clips were selected}
	\label{fig:faceProgress}
\end{figure}
\begin{figure}
	\centering
	\includegraphics[width=0.9\textwidth]{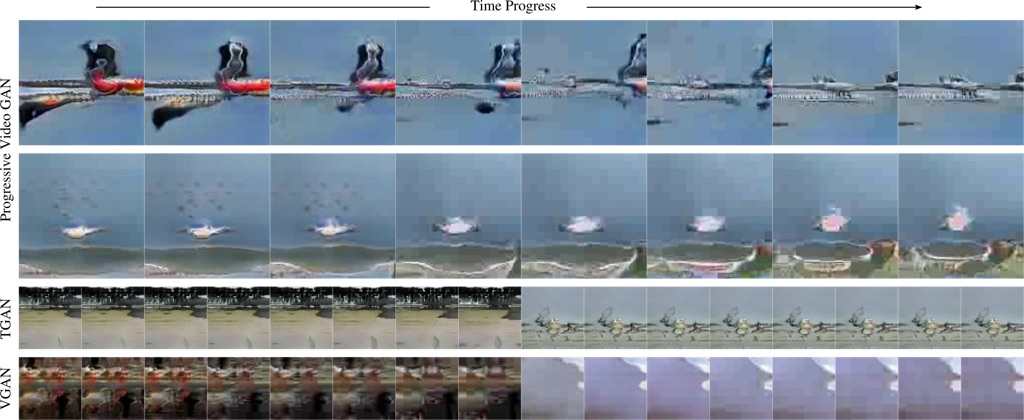}
	\caption{Qualitative comparision of samples from Aeroplane dataset generated by our method with that generated by Video GAN and Temporal GAN.}
	\label{fig:planecomparision}
\end{figure}
\begin{figure}
	\centering
	\includegraphics[width=0.9\textwidth]{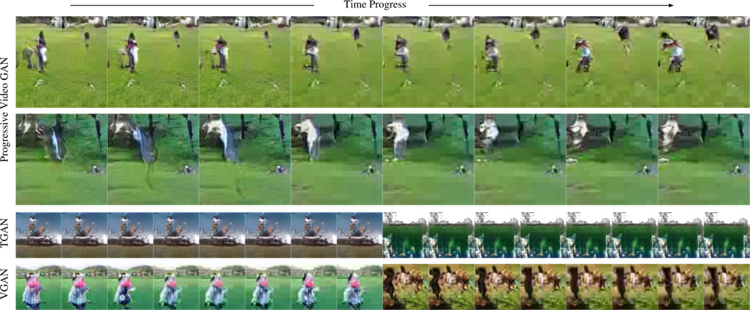}
	\caption{Qualitative comparision of samples from Golf dataset generated by our method with that generated by Video GAN and Temporal GAN.}
	\label{fig:golfcomparision}
\end{figure}
\begin{figure}
	\centering
	\begin{subfigure}
		\centering
		\includegraphics[width=0.45\textwidth]{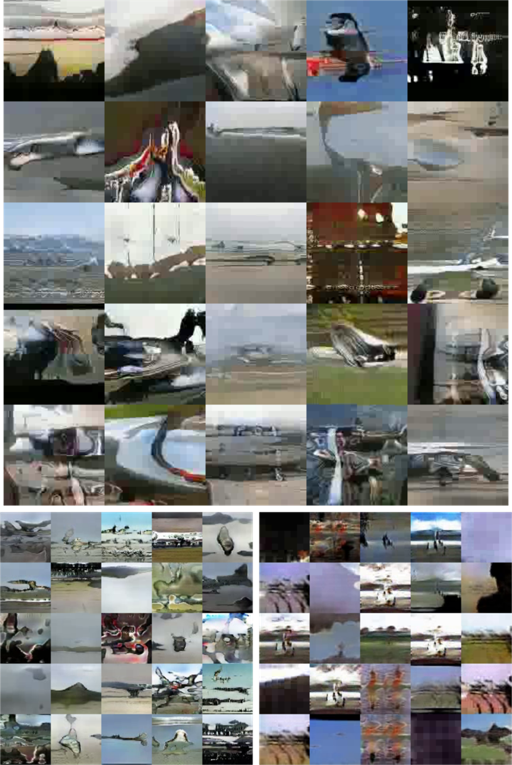}
	\end{subfigure}%
	\begin{subfigure}
		\centering
		\includegraphics[width=0.45\textwidth]{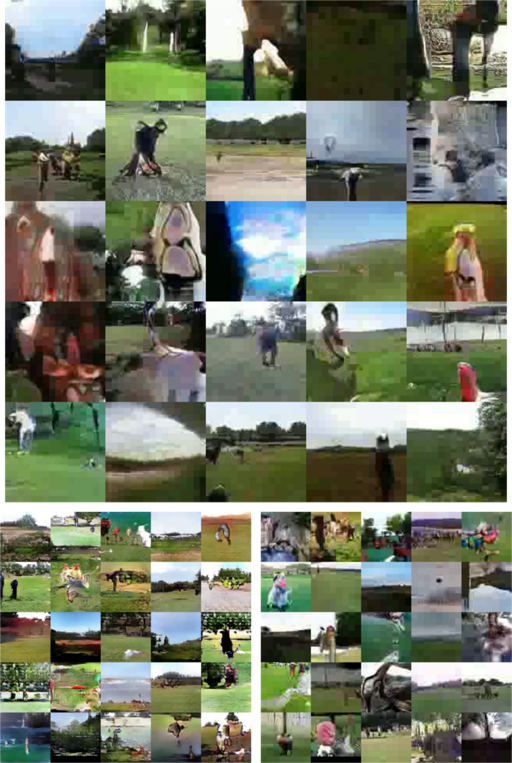}
	\end{subfigure}%
	\caption{Qualitative comparision of clips generated with progressive approach (top), Temporal GAN (bottom left) and Video GAN (bottom right) on aeroplane (left) and golf datasets (right).}
	\label{fig:staticcomparision}	
\end{figure}

As discussed earlier, progressive growing scheme was utilized for training the video generative networks. The improvement in quality and level of details over the course of training is illustrated in Fig.~\ref{fig:faceProgress}. As seen from the figure, more detailed structures appear in the images over the course of training. Furthermore, the generated images look reasonable on TrailerFaces dataset. However, the quality is still not comparable to the quality of generated samples in the case of images \cite{pggan}.
As illustrated in Fig.~\ref{fig:planecomparision}, Fig.~\ref{fig:golfcomparision} and Fig.~\ref{fig:staticcomparision}, the structure of moving objects such aeroplanes, humans and animals is not distinct and they appear as blobs. Though appearance of dynamic objects is not well captured by the network, it can be inferred from Fig.~\ref{fig:planecomparision} and Fig.~\ref{fig:golfcomparision} that temporal dynamics seems more reasonable.

To analyze if the network has overfitted the dataset, we carried out linear interpolation in latent space and generated samples. As seen from Fig.~\ref{fig:golfinterpolation1},\ref{fig:golfinterpolation2},\ref{fig:planeinterpolation1},\ref{fig:planeinterpolation2},\ref{fig:faceinterpolation1},\ref{fig:faceinterpolation2}, samples from all datasets show that our network has good generalization ability and does not overfit the model.


It can be observed that progressive growing technique can generate higher resolution videos without mode collapse or instability issues traditionally suffered by GANs. However, the issue that moving objects appear as blobs in generated samples as reported in \cite{vgan}, is still not completely solved.

\section{Inception score on UCF101}
Same protocol and C3D model proposed in \cite{tgan} was used for computation of inception scores. All scores except our own were taken from \cite{tgan} and \cite{mocogan}. To train our model, central 32 frames from UCF101 were selected and then each frame was resized and cropped to $128\times128$. In our experiments, best Inception score of $13.59$ was obtained with a single, though progressive model. Furthermore, with SWGAN loss, we were able to obtain inception score of $14.56$. Both of these scores are the best result we are aware of.
\setlength{\tabcolsep}{4pt}
\begin{table}[H]
	\begin{center}
		\begin{tabular}{lll}
			\hline\noalign{\smallskip}
			Model  & Inception Score\\
			\noalign{\smallskip}
			\hline
			\noalign{\smallskip}
			VGAN\cite{vgan} & $8.18$ \\
			TGAN\cite{tgan} & $11.85$ \\
			MoCoGAN\cite{mocogan} & $12.42 $ \\	
			Progressive Video GAN & $13.59$ \\
			\textbf{Progressive Video GAN} + \textbf{SWGAN} & \textbf{14.56}\\
			Maximum Possible & $83.18$\\
			\hline
		\end{tabular}
		\caption{Inception scores of Progressive Video GAN compared with other models on UCF101 dataset.}
		\label{table:headings}
	\end{center}
\end{table}
In all cases, first training split of UCF101 dataset was used. However, in \cite{tgan} and \cite{mocogan}, authors randomly sampled $16$ or $32$ consecutive frames during training. In our case, we restricted to central $32$ frames of video during training. 

Surprisingly, inception score started decreasing on training the network further. One possible cause could be smaller minibatch size used at higher resolution. However, further experiment is necessary to make decisive conclusion about the behaviour.

\section{FID}
In this section, we compare FID score of samples generated with our model and one generated with models from VideoGAN and TGAN papers. In C3D model, output of fc-6 layer is 4096-d where as output of pool-5 layer is 8192-d. The output of fc-6 layer of C3D model was used to compute FID score for computational reasons. In order to compute FID score, 10,000 samples were generated with each model. Since VideoGAN and TGAN models were trained to generate on 64x64 resolution videos, we upscaled the videos to $128\times 128$ in order to compute FID score.

\begin{figure}
	\centering
	\begin{subfigure}
		\centering
		\includegraphics[width=0.48\textwidth]{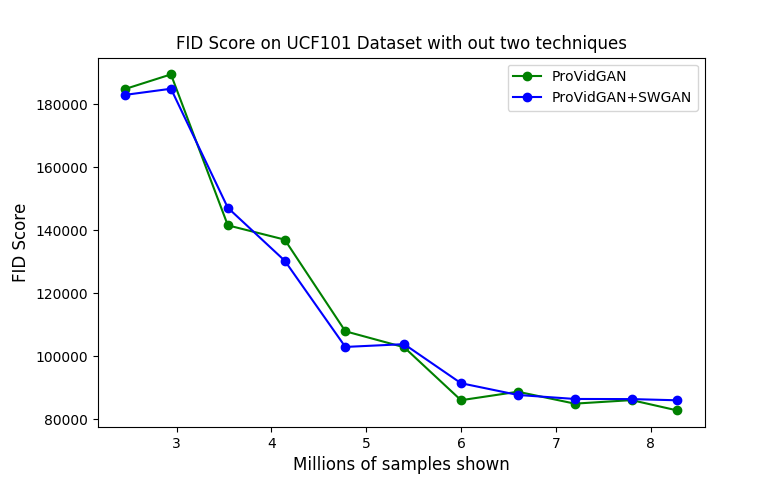}
	\end{subfigure}%
	\begin{subfigure}
		\centering
		\includegraphics[width=0.42\textwidth]{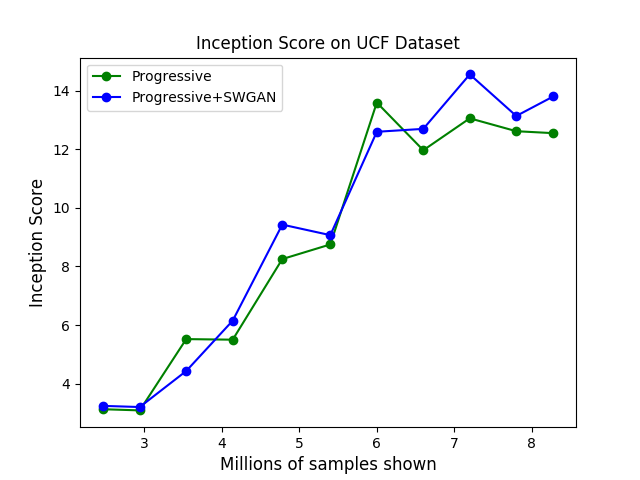}
	\end{subfigure}
	\label{fig:ucfgraphs}
	\caption{Comparison of our models on UCF101 dataset based on FID Score (left) and Inception Score (right).}
\end{figure}

\begin{figure}
	\centering
	\begin{subfigure}
		\centering
		\includegraphics[width=0.48\textwidth]{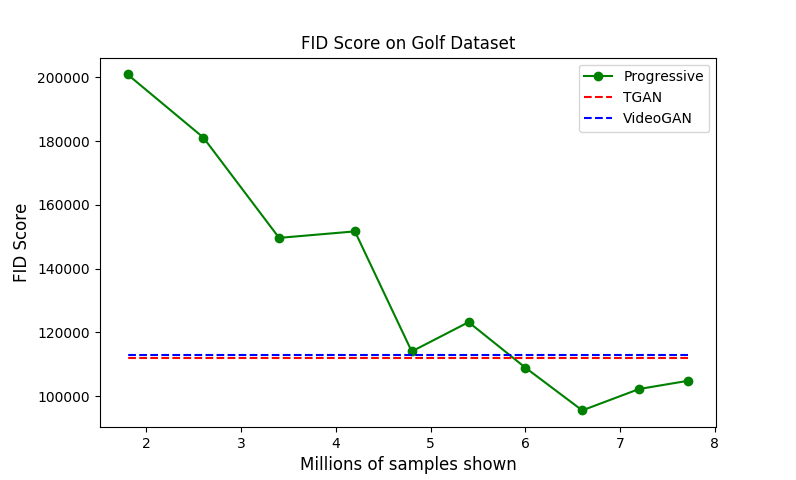}
	\end{subfigure}%
	\begin{subfigure}
		\centering
		\includegraphics[width=0.45\textwidth]{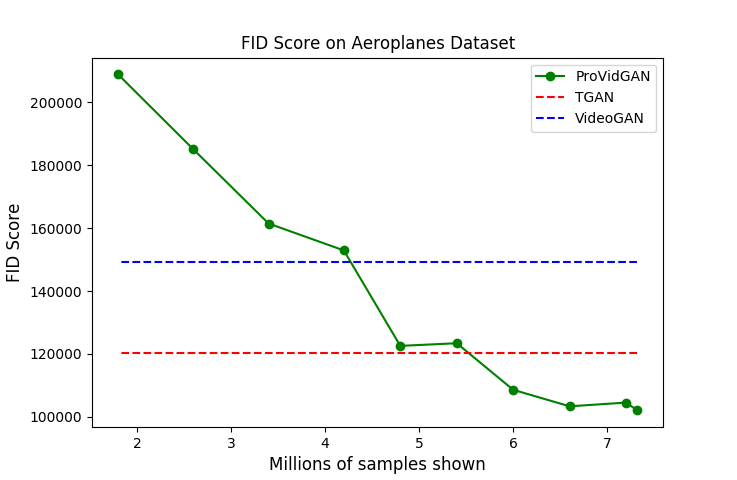}
	\end{subfigure}
	\label{fig:golfplanegraphs}
	\caption{Comparison of our model with TGAN and VideoGAN based on Golf and Aeroplane Datasets as measured by FID score.}
\end{figure}

\begin{table}
	\begin{center}
		\begin{tabular}{lcc}
			\hline\noalign{\smallskip}
			Model  & FID Score on Golf Dataset& FID Score on Aeroplane Dataset\\
			\noalign{\smallskip}
			\hline
			\noalign{\smallskip}
			VGAN\cite{vgan} &  113007 & 149094\\
			TGAN\cite{tgan} & 112029 & 120417\\
			\textbf{Progressive Video GAN} &  \textbf{95544} & \textbf{102049}\\
			\hline
		\end{tabular}
		\caption{Quantitative comparision of Progressive Video GAN with TGAN and VideoGAN based on FID score on Golf and Aeroplane datasets.}
		\label{table:headings}
	\end{center}
\end{table}
To report FID, TGAN and VideoGAN were trained on our own using code available on the internet. It is clear from both datasets that progressive video GAN performs significantly better than TGAN and VideoGAN. The difference is even more prominent in case of Aeroplane dataset where TGAN and Progressive Video GAN perform significantly better than VideoGAN. As mentioned earlier, Golf dataset was stabilized whereas Aeroplane dataset was not. This is easily explained by the fact that VideoGAN assumes stable background whereas TGAN and Progressive Video GAN make no such assumptions.

%

\chapter{Conclusion and Discussion}
In this work, we explored the use of progressive growing of GANs for video generation. By progressively growing the network, we were able to generate videos of up to 256x256 resolution and 32 frames. Our model performed better than existing state-of-art models on UCF101 dataset as measured by Inception Score and FID Score. We obtained state of art inception score of 14.56 on UCF101 dataset. This is significant improvement over state-of-art score reported in \cite{mocogan}. This shows that the idea of first training on simpler problem and progressively increasing the complexity of the problem is also effective for video generative models. Though we observed that use of Sliced Wasserstein metric as loss for training GANs improves performance of the model as measured by inception score in some cases, more experiment is needed to make a definite conclusion. Perceptually, the results obtained are still not as realistic as were obtained in the case of image based on methods. Recently there has been surge in models that use optical flow\cite{ftgan}\cite{timelapse}\cite{texture} to generate videos. This is a promising direction to explore.

Following the approach of \cite{pggan}, the discriminator we used was mirror reflection of the generator architecture. Due to this, the network architecture of discriminator in our model is drastically different than that of C3D model which was used for quantitative comparison. As there are fewer feature channels in our Discriminator, it is weaker than C3D model or the discriminators used for previous works\cite{tgan}\cite{mocogan}\cite{vgan}. It is interesting to note that our model performs better than existing approaches despite a discriminator with fewer parameters.

There is enough space to explore if incorporation of the idea of progressive growing of gans to more complex architectures such as \cite{tgan}\cite{mocogan} that have different temporal and spatial generators. Furthermore, incorporation of prior information such as class label can be expected to drastically improve the results. The higher quality image generation in \cite{pggan} was possible due to higher quality well aligned training data. Application of similar super-resolution techniques can be expected to further improve the quality of video generation. Furthermore, well alignment of faces in TrailerFaces dataset can be expected to significantly impact the quality of videos generated. However, the problem of alignment of faces in videos is not as well-posed problem as alignment of faces in images and is more challenging.

For evaluation of our generative models, we relied on C3D models trained on Sports-1M dataset and fine tuned in UCF101 dataset. This model has clip level accuracy of approximately $75\%$ on test split. The effectiveness of a model with far from $100\%$ accuracy for evaluation of generative models is a research problem in itself. Furthermore, recently accuracy of $98.0\%$ was achieved on UCF101 with Two-Stream Inflated 3D ConvNet (I3D)\cite{quovadis}. We can expect the I3D model to give more valid Inception and FID Scores. Furthermore, for training on UCF101, we only used central 32 frames of UCF101 for training. This method was slightly different than use of randomly selected 32 consecutive frames\cite{tgan}\cite{mocogan}. It is also important to point out that, we obtained the best possible inception score of $83.18\%$. Whereas, in \cite{tgan}, best possible score of $34.33\%$ was reported. It is possible that authors downsampled original videos to 64x64 and then again upscaled to 128x128 before feeding into C3D model. This is reasonable as their model is trained on 64x64 videos. We directly downscaled to 128x128 before feeding into the network as we designed our network to train on 128x128 resolution.


\appendix
%
\chapter{Network Architecture}
\begin{table}
	\begin{center}
		\begin{tabular}{lccc}
			\hline\noalign{\smallskip}
			Generator  & Activation & Output shape & Parameters\\
			\noalign{\smallskip}
			\hline
			\noalign{\smallskip}
			Latent vector &  -& 128x1x1x1&-\\
			Fully-connected & LReLU& 8192x1x1x1 &1.04m\\
			Conv 3x3x3 & LReLU& 128x4x4x4&0.44m\\\hline						
			Upsample & - & 128x8x8x8&-\\
			Conv 3x3x3 & LReLU& 128x8x8x8&0.44m\\
			Conv 3x3x3 & LReLU& 128x8x8x8&0.44m\\\hline						
			Upsample & - & 128x8x16x16&-\\
			Conv 3x3x3 & LReLU& 128x8x16x16&0.44m\\
			Conv 3x3x3 & LReLU& 128x8x16x16&0.44m\\\hline						
			Upsample & - & 128x8x32x32&-\\
			Conv 3x3x3 & LReLU& 64x8x32x32&0.22m\\
			Conv 3x3x3 & LReLU& 64x8x32x32&0.22m\\\hline						
			Upsample & - & 64x16x64x64&-\\
			Conv 3x3x3 & LReLU& 32x16x64x64&55k\\
			Conv 3x3x3 & LReLU& 32x16x64x64&27k\\\hline						
			Upsample & - & 32x16x128x128&-\\
			Conv 3x3x3 & LReLU& 16x16x128x128&13.8k\\
			Conv 3x3x3 & LReLU& 16x16x128x128&6.9k\\\hline
			Upsample & - & 16x32x256x256&-\\
			Conv 3x3x3 & LReLU& 8x32x256x256&3.4k\\
			Conv 3x3x3 & LReLU& 8x32x256x256&1.7k\\\hline
			Conv 1x1x1 & LReLU& 3x32x256x256&27\\\hline			
			Total Parameters & & & 3.7m \\ \hline
		\end{tabular}
		\caption{Generator architecture for generation of 256x256x32 videos.}
		\label{table:genarch}
	\end{center}
\end{table}

\begin{table}
	\begin{center}
		\begin{tabular}{lccc}
			\hline\noalign{\smallskip}
			Discriminator  & Activation & Output shape & Parameters\\
			\noalign{\smallskip}
			\hline
			\noalign{\smallskip}
			Input Image &  -& 128x1x1&-\\
			Conv 1x1x1 & LReLU& 128x4x4x4&32\\
			Conv 3x3x3 & LReLU& 128x4x4x4&1.73k\\
			Conv 3x3x3 & LReLU& 128x4x4x4&3.47k\\
			Downsample & - & 128x8x8x8&-\\\hline
			Conv 3x3x3 & LReLU& 128x8x8x8&6.92k\\
			Conv 3x3x3 & LReLU& 128x8x8x8&13.85k\\
			Downsample & - & 128x8x16x16&-\\\hline
			Conv 3x3x3 & LReLU& 128x8x16x16&27.68k\\
			Conv 3x3x3 & LReLU& 128x8x16x16&55.36k\\
			Downsample & - & 128x8x32x32&-\\\hline
			Conv 3x3x3 & LReLU& 64x8x32x32&0.11m\\
			Conv 3x3x3 & LReLU& 64x8x32x32&0.22m\\
			Downsample & - & 64x16x64x64&-\\\hline
			Conv 3x3x3 & LReLU& 32x16x64x64&0.44k\\
			Conv 3x3x3 & LReLU& 32x16x64x64&0.44k\\
			Downsample & - & 32x16x128x128&-\\\hline
			Conv 3x3x3 & LReLU& 16x16x128x128&0.44m\\
			Conv 3x3x3 & LReLU& 16x16x128x128&0.44m\\
			Downsample & - & 16x32x256x256&-\\\hline
			Minibatch Stddev&-&129x4x4x4&- \\
			Conv 3x3x3 & LReLU& 8x32x256x256&.44m\\
			Fully-connected & linear & 1x1x1x128&1.04m\\
			Fully-connected & linear & 1x1x1x1&129\\
			Total Parameters & & & 3.7m \\\hline
		\end{tabular}
		\caption{Discriminator architecture for generation of 256x256x32 videos.}
		\label{table:discarch}
	\end{center}
\end{table}

\chapter{Latent Space Interpolations}
\section*{Golf Dataset}
\begin{figure}[H]
	\centering
	\includegraphics[width=.7\textwidth]{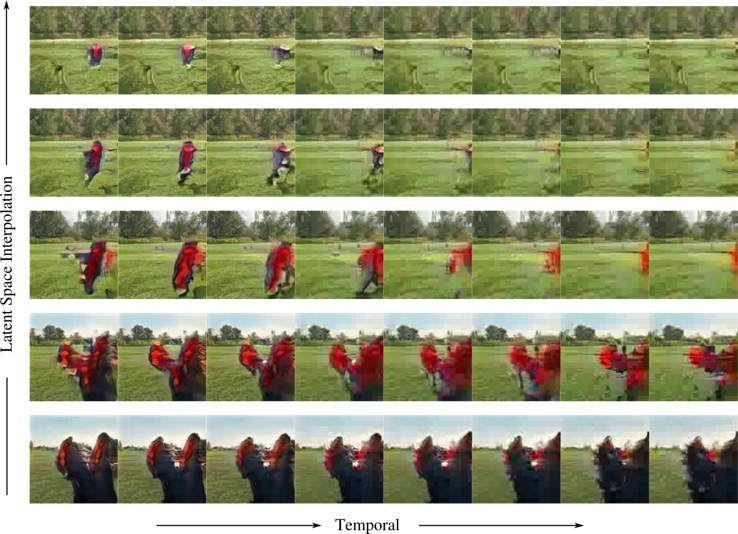}
	\caption{Linear interpolation in latent space to generate samples from Golf dataset - 1.}
	\label{fig:golfinterpolation1}	
\end{figure}
\begin{figure}[H]
	\centering
	\includegraphics[width=.7\textwidth]{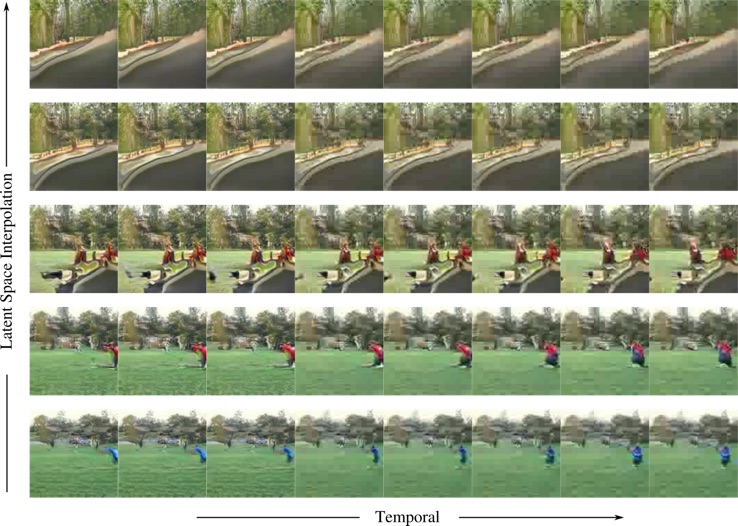}
	\caption{Linear interpolation in latent space to generate samples from Golf dataset - 2}
	\label{fig:golfinterpolation2}	
\end{figure}

\section*{Aeroplane Dataset}
\begin{figure}[H]
	\centering
	\includegraphics[width=.7\textwidth]{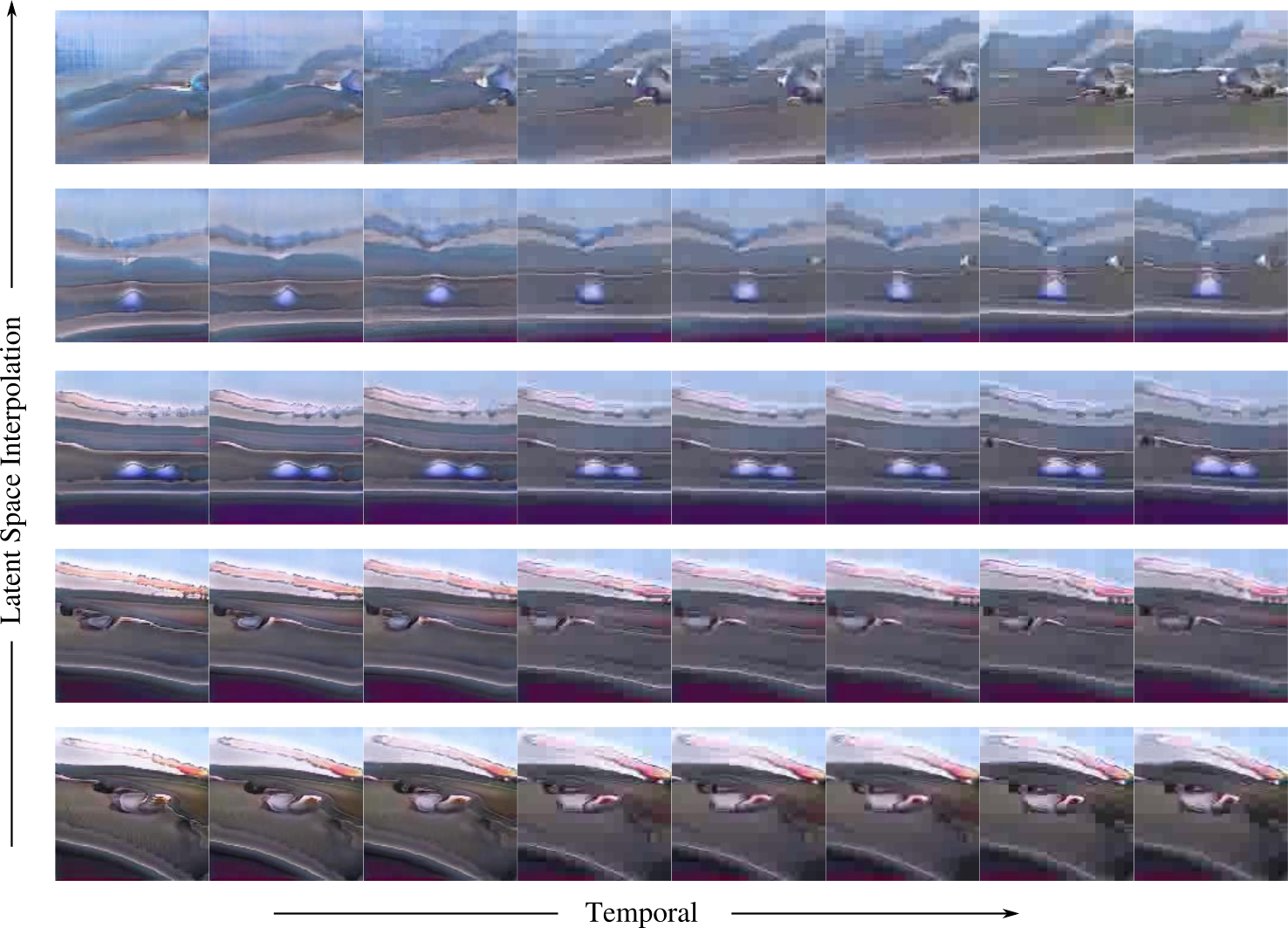}
	\caption{Linear interpolation in latent space to generate samples from Aeroplane dataset - 1}
	\label{fig:planeinterpolation1}	
\end{figure}

\begin{figure}[H]
	\centering
	\includegraphics[width=.7\textwidth]{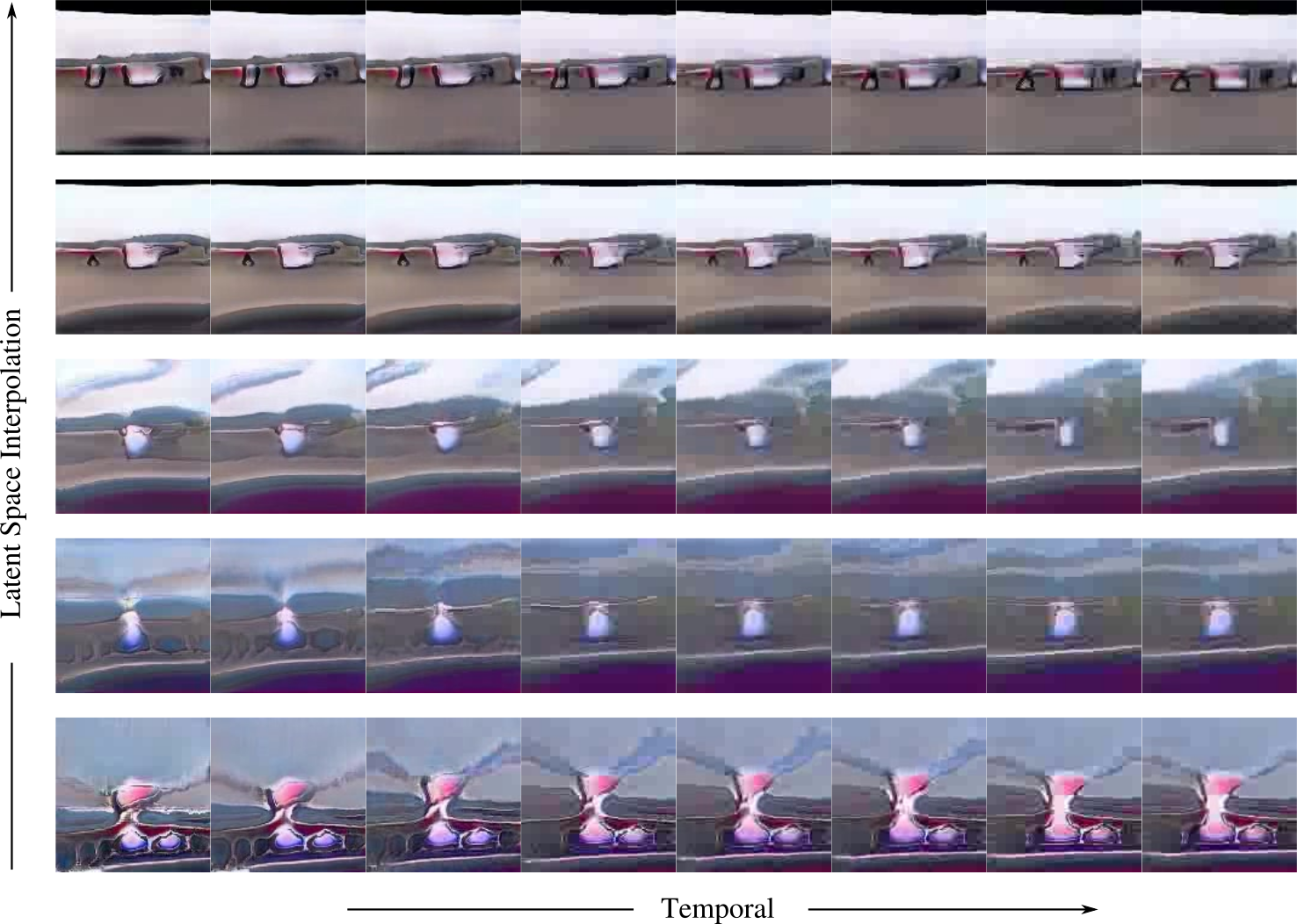}
	\caption{Linear interpolation in latent space to generate samples from Aeroplane dataset - 2}
	\label{fig:planeinterpolation2}	
\end{figure}

\section*{TrailerFaces}
\begin{figure}[H]
	\centering
	\includegraphics[width=.7\textwidth]{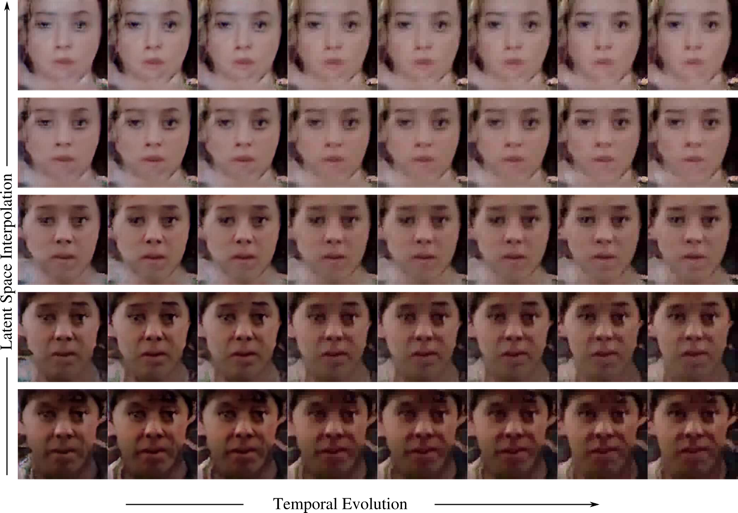}
	\caption{Linear interpolation in latent space to generate samples from TrailerFaces dataset - 1}
	\label{fig:faceinterpolation1}	
\end{figure}

\begin{figure}[H]
	\centering
	\includegraphics[width=.7\textwidth]{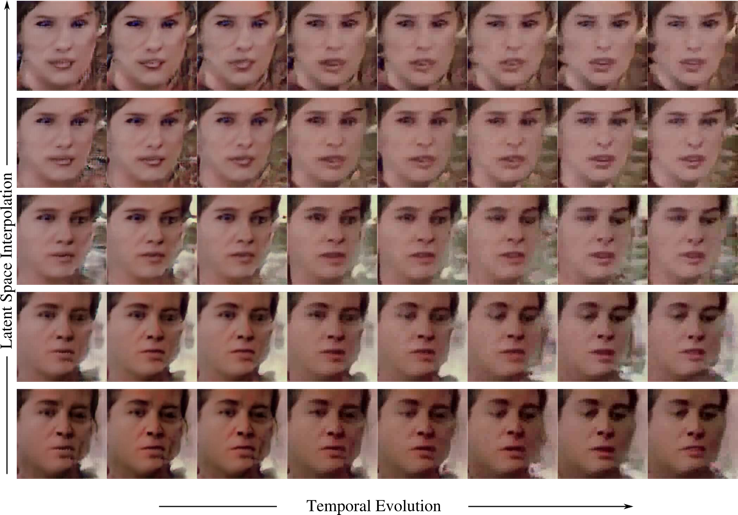}
	\caption{Linear interpolation in latent space to generate samples from TrailerFaces dataset - 2}
	\label{fig:faceinterpolation2}	
\end{figure}

%
%

\bibliographystyle{plain}

\end{document}